\documentclass[acmtog,nonacm]{acmart}
\acmSubmissionID{1234}

\usepackage{booktabs} 

\usepackage[ruled]{algorithm2e} 

\SetAlFnt{\small}
\SetAlCapFnt{\small}
\SetAlCapNameFnt{\small}
\SetAlCapHSkip{0pt}

\acmJournal{TOG}

\usepackage{subcaption}
\usepackage{wrapfig}

\usepackage{multirow}
\usepackage{mathtools}
\usepackage{bbm}

\newif\ifdraft
\draftfalse
\ifdraft
\newcommand{\KCW}[1]{{\color{blue}[\textbf{Jackson:} #1]}}
\newcommand{\kac}[1]{{\color{purple}[\textbf{Kfir} #1]}}
\newcommand{\stc}[1]{{\color{red}[\textbf{Sergey:} #1]}}
\newcommand{\yfc}[1]{{\color{orange}[\textbf{Yuwei:} #1]}}
\newcommand{\dosc}[1]{{\color{teal}[\textbf{Daniil:} #1]}}
\newcommand{\kcwqa}[1]{{\color{blue}#1}}
\newcommand{\ka}[1]{{\color{purple}#1}}

\newcommand{\dos}[1]{{\color{teal}#1}}
\newcommand{\todo}[1]{{\color{blue}[TODO: #1]}}

\else
\newcommand{\KCW}[1]{}
\newcommand{\kac}[1]{}
\newcommand{\stc}[1]{}
\newcommand{\yfc}[1]{}
\newcommand{\todo}[1]{}
\newcommand{\kcwqa}[1]{{\color{black}#1}}
\newcommand{\ka}[1]{{\color{black}#1}}

\newcommand{\dos}[1]{}
\newcommand{\dosc}[1]{}
\fi

\usepackage[capitalize]{cleveref}
\crefname{equation}{Eqn.}{Eqns.}
\Crefname{equation}{Equation}{Equations}
\crefname{section}{Sec.}{Secs.}
\Crefname{section}{Section}{Sections}
\crefname{table}{Tab.}{Tabs.}
\Crefname{table}{Table}{Tables}

\begin{document}
\title{MoA: Mixture-of-Attention for Subject-Context Disentanglement in Personalized Image Generation}

\author{Kuan-Chieh (Jackson) Wang}
\affiliation{%
 \institution{Snap Inc.}
 \country{USA}
 }
\email{jwang23@snapchat.com}
\author{Daniil Ostashev}
\affiliation{%
 \institution{Snap Inc.}
 \country{UK}
}
\author{Yuwei Fang}
\affiliation{%
 \institution{Snap Inc.}
 \country{USA}
}
\author{Sergey Tulyakov}
\affiliation{%
 \institution{Snap Inc.}
 \country{USA}
}
\author{Kfir Aberman}
\affiliation{%
 \institution{Snap Inc.}
 \country{USA}
}
\email{kaberman@snapchat.com}

\begin{abstract}
We introduce a new architecture for personalization of text-to-image diffusion models, coined Mixture-of-Attention (MoA). Inspired by the Mixture-of-Experts mechanism utilized in large language models (LLMs), MoA distributes the generation workload between two attention pathways: a personalized branch and a non-personalized prior branch.
MoA is designed to retain the original model's prior by fixing its attention layers in the prior branch, while minimally intervening in the generation process with the personalized branch that learns to embed subjects in the layout and context generated by the prior branch.
A novel routing mechanism manages the distribution of pixels in each layer across these branches to optimize the blend of personalized and generic content creation. 
Once trained, MoA facilitates the creation of high-quality, personalized images featuring multiple subjects with compositions and interactions as diverse as those generated by the original model.
Crucially, MoA enhances the distinction between the model's pre-existing capability and the newly augmented personalized intervention, thereby offering a more disentangled subject-context control that was previously unattainable.  Project page: \url{https://snap-research.github.io/mixture-of-attention}.

\end{abstract}

\begin{teaserfigure}
\includegraphics[width=\textwidth]{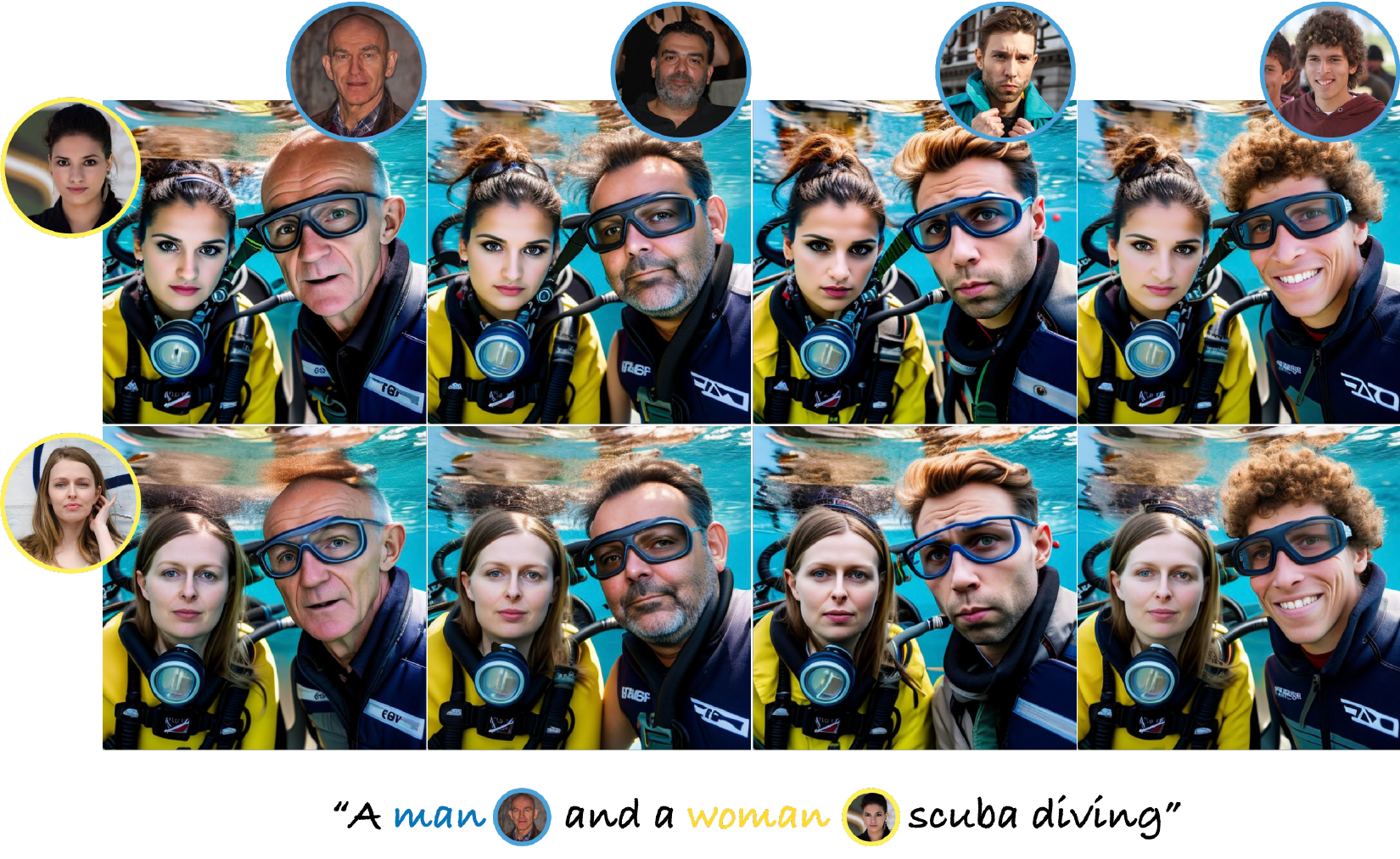}
  \caption{\textbf{Mixture-of-Attention (MoA)} architecture enables multi-subject personalized generation with \textbf{subject-context disentanglement}. Given a multi-modal prompt that includes text and input images of human subjects, our model can generate the subjects in a fixed context and composition, without any predefined layout. MoA minimizes the intervention of the personalized part in the generation process, enabling the decoupling between the model's pre-existing capability and the personalized portion of the generation.
  }
\label{fig:1}
\end{teaserfigure}

%
%





\maketitle

\section{Introduction}
\label{sec:intro}


Recent progress in AI-generated visual content has been nothing short of revolutionary, fundamentally altering the landscape of digital media creation. Foundation models have democratized the creation of high-quality visual content, allowing even novice users to generate impressive images from simple text prompts~\cite{rombach2022high,saharia2022photorealistic,ramesh2022hierarchical}. Among the myriad avenues of research within this field, personalization stands out as a crucial frontier.  
It aims at tailoring the output of a generative model to include user-specific subjects with high fidelity, thereby producing outputs that resonate more closely with individual assets or preferences~\cite{ruiz2023dreambooth,gal2022image}. While being able to say "Create a photo of people scuba diving!" is fun, the experience becomes personal and fosters a stronger emotional connection when one can say "Create a photo of \emph{me and my friend} scuba diving!" (see~\cref{fig:1}). 



Despite the remarkable generative capabilities of these models, current personalization techniques often falter in preserving the richness of the original model. Herein, we refer to the model before personalization as the prior model. In finetuning-based personalization techniques, due to the modifications of the weights, the model tends to overfit to certain attributes in the distribution of the input images (e.g., posture and composition of subjects) or struggles to adhere adequately to the input prompt.
This issue is exacerbated when it comes to multiple subjects; the personalized model struggles to generate compositions and interactions between the subjects that otherwise appear within the distribution of the non-personalized model. 
Even approaches that were optimized for multi-subject generation modify the original model's weights, resulting in compositions that lack diversity and naturalness \cite{xiao2023fastcomposer,po2023orthogonal}.
Hence, it is advisable to purse a personalization method that is \emph{prior preserving}. Functionally, we refer to a method as \emph{prior preserving} if the model retains its responsiveness to changes in the text-prompt and random seed like it does in the prior model.




A good personalization method should address the aforementioned issues.
In addition, it should allow the creation process to be \emph{spontaneous}. 
Namely, iterating over ideas should be fast and easy.
Specifically, our requirements are summarized by the following: 
\begin{enumerate}

\item  \textit{Prior preserving}.  The personalized model should retain the ability to compose different elements, and be faithful to the \emph{interaction} described in the text prompt like in the prior model.  Also, the distribution of images should be as diverse as in the prior model.




\item \textit{Fast generation}. The generation should be fast to allow the users to quickly iterate over many ideas.  Technically, the personalized generation process should be inference-based and should not require optimization when given a new subject.

\item \textit{Layout-free}. Users are not required to provide additional layout controls (e.g. segmentation mask, bounding box, or human pose) to generate images. Requiring additional layout control could hinder the creative process, and restrict the diversity of the distribution.


\end{enumerate}

\begin{figure}[t]
  \centering
  \includegraphics[width=\columnwidth]{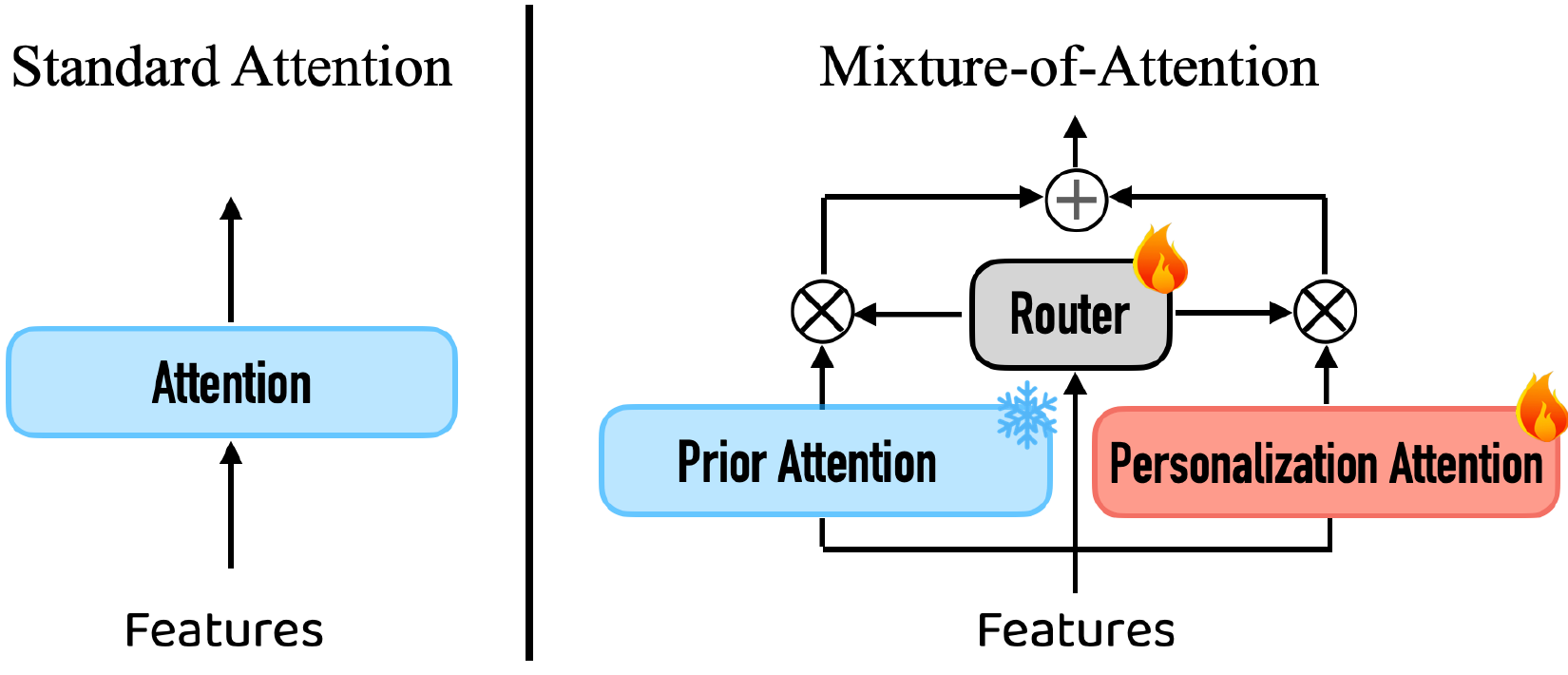}
  \caption{ \textbf{Mixture-of-Attention.} Unlike the standard attention mechanism (left), MoA is a dual attention pathways that contains a trainable personalized attention branch and a non-personalized fixed attention branch that is copied from the original model (prior attention). In addition, a routing mechanism manages the distribution of pixels in each layer across these branches to optimize the blend of personalized and generic content creation. 
  }
  \label{fig:system}
\end{figure}

To achieve these goals, we introduce Mixture-of-Attention (MoA) (see~\cref{fig:system}).
Inspired by the Mixture-of-Expert (MoE) layer~\cite{jacobs1991adaptive} and its recent success in scaling language models~\cite{roller2021hash}, MoA extends the vanilla attention mechanism into multiple attention blocks (i.e. experts), and has a router network that softly combines the different experts. In our case, MoA distributes the generation between personalized and non-personalized attention pathways. It is designed to retain the original model's prior by fixing its attention layers in the prior (non-personalized) branch, while minimally intervening in the generation process with the personalized branch. The latter learns to embed subjects that are depicted in input images, via encoded visual tokens that is injected to the layout and context generated by the prior branch. This mechanism is enabled thanks to the router that blends the outputs of the personalized branch only at the subject pixels (i.e. foreground), by learning soft segmentation maps that dictate the distribution of the workload between the two branches. This mechanism frees us from the trade-off between identity preservation and prompt consistency. 

Since MoA distinguishes between the model's inherent capabilities and the personalized interventions, it unlocks new levels of disentangled control in personalized generative models (as demonstrated in Fig.~\ref{fig:1}). This enables us to create various applications with MoA such as subject swap, subject morphing, style transfer, etc,that was previously challenging to attain.
In addition, due to the existence of the fixed prior branch, MoA is compatible with many other diffusion-based image generation and editing techniques, such as ControlNet~\cite{zhang2023adding} or inversion techniques that unlocks a novel approach to easily replace subjects in a real images (see~\cref{sec:app}).

\section{Related Works}


\subsection{Personalized Generation}
Given the rapid progress in foundation text-conditioned image synthesis with diffusion models ~\cite{Ho2020DenoisingDP, Song2020DenoisingDI, Dhariwal2021DiffusionMB, Ho2022ClassifierFreeDG, Rombach2021HighResolutionIS, Pandey2022DiffuseVAEEC, nichol2021improved}, \emph{personalized} generation focuses on adapting and contextualizing the generation to a set of desired subject using limited input images, while retaining the powerful generative capabilities of the foundation model.
Textual Inversion (TI)~\cite{gal2022image} addresses  the personalization challenge by utilizing a set of images depicting the same subject to learn a special text token that encodes the subject.  
Yet, using only the input text embeddings is limited in expressivity.
Subsequent research, such as $\mathcal{P}+$~\cite{voynov2023p+} and NeTI~\cite{alaluf2023neural}, enhance TI with a more expressive token representation, thereby refining the alignment and fidelity of generated subjects.
DreamBooth (DB)~\cite{ruiz2023dreambooth} can achieve much higher subject fidelity by finetuning the model parameters. 
E4T~\cite{gal2023encoder} introduced a pretrained image encoder that jump starts the optimization with image features extracted from the subject image, and is able to substantially reduce the number of optimization steps.
Other extensions include multi-subject generation~\cite{custom-diffusion}, generic objects~\cite{li2024blipdiffusion}, human-object composition~\cite{liu2023cones1,liu2023cones2}, subject editing~\cite{tewel2023key}, improving efficiency~\cite{han2023svdiff,dblora,hu2022lora}, and using hypernetworks~\cite{arar2023domain,ruiz2023hyperdreambooth}.
These approaches fall under the \emph{optimization-based} category where given a new subject, some parameters of the model are to be optimized. 
Because of the optimization which modifies the original parameters of the model, these methods are inevitably slow and prone to breaking prior preservation.
In contrast, MoA falls in the \emph{optimization-free} category.  
These approaches do not require optimization when given a new subject.  
They augment the foundation T2I model with an image encoder and finetune the augmented model to receive image inputs. 
Relevant methods in this category include IP-Adapter~\cite{ye2023ip-adapter} and InstantID~\citep{wang2024instantid}.
A critical difference is, in MoA, the image features are combined with a text token (e.g. `man') and processed by the cross attention layer in the way that the cross attention layer was trained.  In IP-Adapter and InstantID, the image features are combined with the output of attention layers and do not have binding to a specific text token.  
This design makes it hard to leverage the native text understanding and text-to-image composition of the pretrained T2I model.
It also makes combining multiple image inputs challenging.  
For similar reasons, other optimization-free approaches that focus on the single subject setting include ELITE~\cite{wei2023elite}, InstantBooth~\cite{shi2023instantbooth}, PhotoMaker~\cite{li2023photomaker}, LCM-Lookahead~\cite{gal2024lcm}.
A remedy is to introduce layout controls and mask the different image inputs in the latent space, but this led to rigid outputs and a brittle solution.
In stark contrast, since MoA injects the image inputs in the text space, injecting multiple input images is trivial.  
In addition, by explicitly having a prior branch, MoA preserves the powerful text-to-image capabilities of the prior foundation model.

\subsection{Multi-subject Generation}
\begin{figure}[t]
  \centering

  \includegraphics[width=\columnwidth]{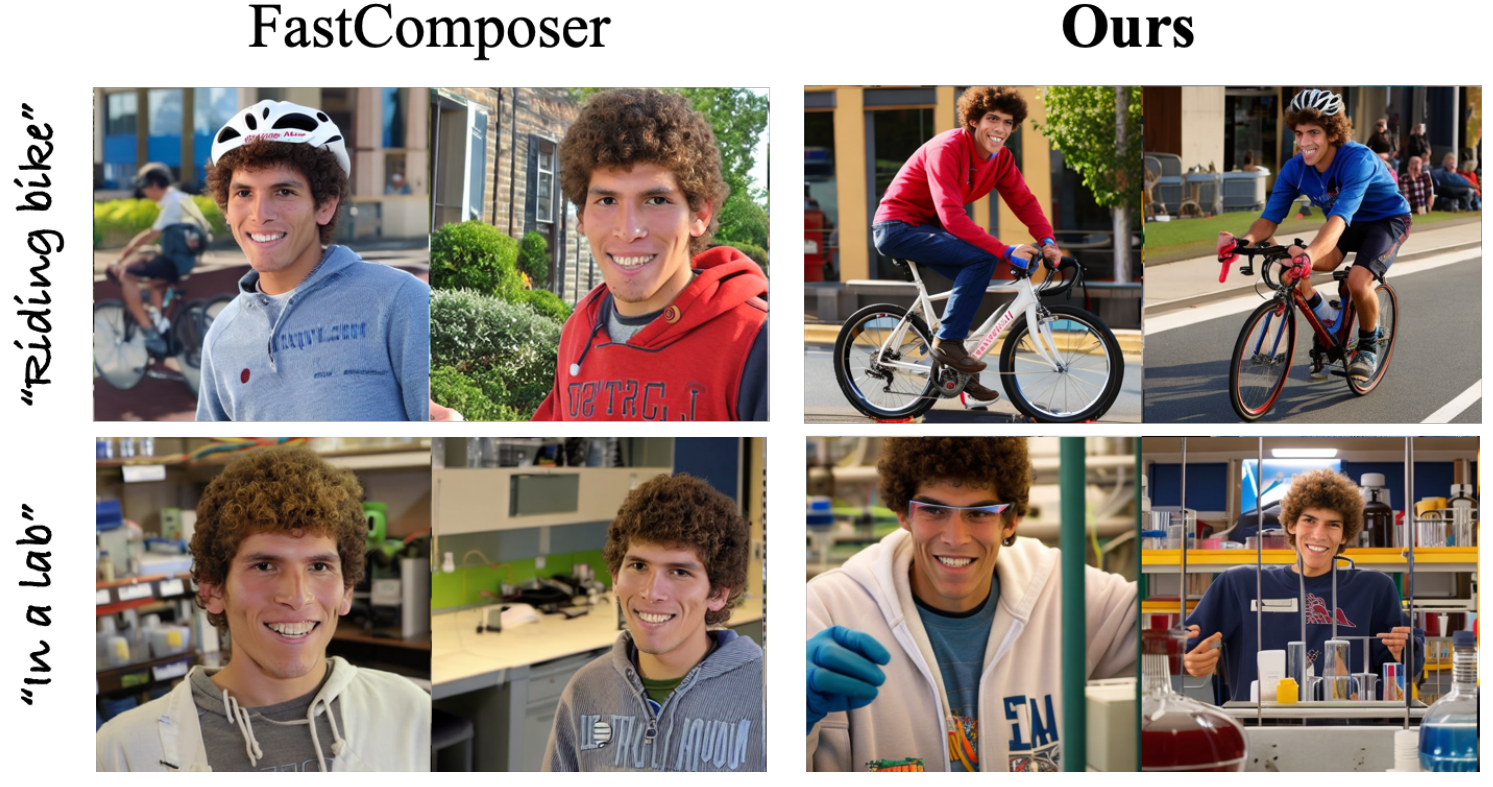}
  
  \caption{ \textbf{Comparing image variations.} In contrast to Fastcomposer~\cite{xiao2023fastcomposer}, our method (MoA) is able to generate images with diverse compositions, and foster interaction of the subject with what is described in the text prompt. }
  \label{fig:fastcomposer-comparison}
\end{figure}

Extending personalized generation to the multi-subject setting is not trivial. 
Naive integration of multiple subjects often leads to issues like a missing subject, poor layout, or subject interference (a.k.a. identity leak) where the output subjects looks like a blended version of the inputs.
Custom Diffusion~\cite{kumari2022customdiffusion} and Modular Customization~\cite{po2023orthogonal} proposed ways to combine multiple DB parameters without the subjects interfering with each other using constrained optimization techniques.
Mix-of-show~\cite{gu2024mix} proposed regionally controllable sampling where user specified bounding boxes are used to guide the generation process. 
InstantID~\cite{wang2024instantid} can also achieve multi-subject generation using bounding boxes as additional user control.  
The idea of using bounding box or segmentation mask to control the generation process has be used in other settings~\cite{avrahami2023break,hertz2022prompt,bar2023multidiffusion}.
In addition to burdening the users to provide layout, methods that require region control naturally results in images that appear more rigid.  The subjects are separated by their respective bounding boxes, and lack interaction.  
In contrast, while MoA can work with additional layout condition, it does not require such inputs just like the prior T2I model does not require layout guidance. 
Fastcomposer~\cite{xiao2023fastcomposer} is the closest method as it also injects the subject image as text tokens and handles multiple subjects without layout control.
Except, generated images from Fastcomposer have a characteristic layout of the subjects and lack subject-context interaction, which indicates the lack of prior preservation (see~\cref{fig:fastcomposer-comparison}).  Since Fastcomposer finetunes the base model's parameters, it inevitably deviates away from the prior model and has to trade-off between faithful subject injection and prior preservation. 
MoA is free from this trade-off thanks to the dual pathways and a learned router to combine both the frozen prior expert and learned personalization expert.

\section{Method}

\begin{figure*}[t]
  \centering
  \includegraphics[width=\textwidth]{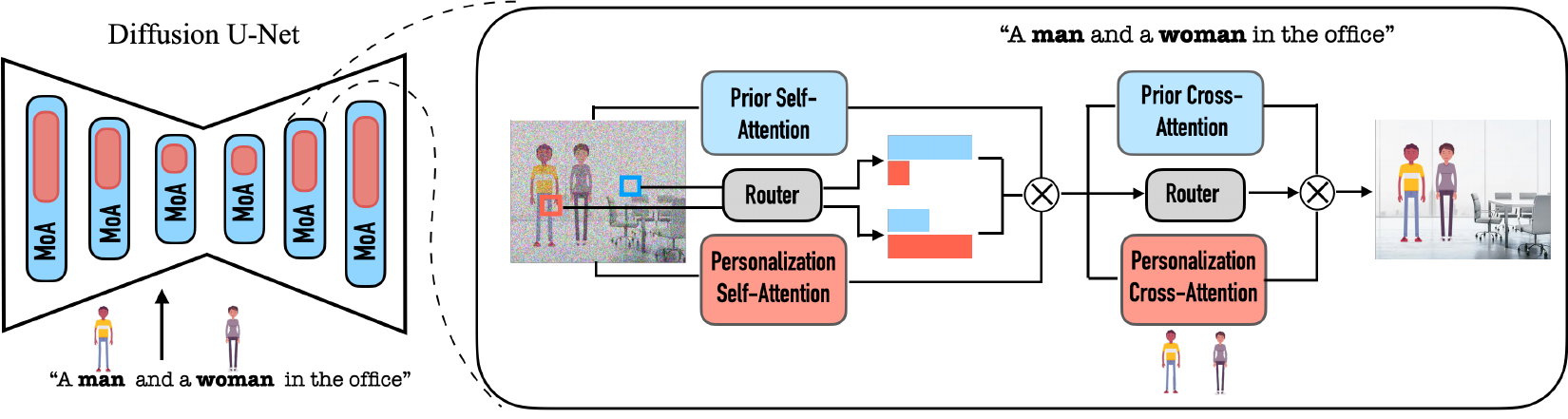}
  \caption{ \textbf{Text-to-Image Diffusion Models with MoA.} Our architecture expands the original diffusion U-Net by replacing each attention block (self and cross) with MoA. In each inference step, a MoA block receives the input image features and passes them to the router, which decides how to balance the weights between the output of the personalized attention and the output of the original attention block. Note that the images of the subjects are injected only through the personalized attention branch; hence, during training, where the router is encouraged to prioritize the prior branch, the result is that only the minimal necessary information required for generating the subjects will be transferred to the personalized attention.
  }
  \label{fig:unet}
\end{figure*}

In this section, we introduce the \emph{Mixture-of-Attention} (MoA) layer (see~\cref{fig:system}) and explain how it can be integrated into text-to-image (T2I) diffusion models for subject-driven generation.
In its general form, a MoA layer has multiple attention layers, each with its own projection parameters and a router network that softly combines their outputs.
In this work, we use a specific instantiation suitable for personalization, which contains two branches: a fixed ``prior'' branch that is copied from the original network, a trainable ``personalized'' branch that is finetuned to handle image inputs, and a router trained to utilize the two experts for their distinct functionalities. 
MoA layer is used in-place of all attention layers in a pretrained diffusion U-Net (see~\cref{fig:unet}).
This architecture enables us to augment the T2I model with the ability to perform subject-driven generation with disentangled subject-context control, thereby preserving the diverse image distribution inherent in the prior model.

\subsection{Background}
\paragraph{Attention Layer}
An attention layer first computes the attention map using query, $\mathbf{Q} \in \mathbb{R}^{l_q \times d}$, and key, $\mathbf{K} \in \mathbb{R}^{l_k \times d}$ where $d$ is the hidden dimension and $l_q, l_k$ are the numbers of the query and key tokens, respectively.   
The attention map is then applied to the value, $\mathbf{V} \in \mathbb{R}^{l \times d}$.  The attention operation is described as follows: 
\begin{gather}
    \mathbf{Z'}=\text{Attention}(\mathbf{Q},\mathbf{K},\mathbf{V}) = \text{Softmax}(\frac{\mathbf{Q}\mathbf{K}^{\top}}{\sqrt{d}})\mathbf{V},\\
    \label{eqn:attn}
    \mathbf{Q} = \mathbf{Z}\mathbf{W}_q, \;\; \mathbf{K} = \mathbf{C}\mathbf{W}_k,  \;\; \mathbf{V} = \mathbf{C}\mathbf{W}_v, 
\end{gather}
where $\mathbf{W}_q \in \mathbb{R}^{d_z \times d}, \mathbf{W}_k \in \mathbb{R}^{d_c \times d}, \mathbf{W}_v \in \mathbb{R}^{d_c \times d}$ are the projection matrices of the attention operation that map the different inputs to the same hidden dimension, $d$.  $\mathbf{Z}$ is the hidden state and $\mathbf{C}$ is the condition.  In self attention layers, the condition is the hidden state, $\mathbf{C} = \mathbf{Z}$.  In cross attention layers of T2I models, the condition is the text conditioning $\mathbf{C} = \mathbf{T}$.

\paragraph{Diffusion U-Net}
At the core of a T2I diffusion model lies a U-Net which consists of a sequence of transformer blocks, each with self attention and cross attention layers.  
Each attention layer has its own parameters.  
In addition, the U-Net is conditioned on the diffusion timestep. 
Putting it together, the input to the specific attention layer is dependent on the U-Net layer $l$ and the diffusion timestep $t$:
\begin{equation}
    \mathbf{Q}^{t,l} = \mathbf{Z}^{t,l}\mathbf{W}_q^{l},  \;\; \mathbf{K}^{t,l} = \mathbf{C}^{t,l}\mathbf{W}_k^{l},  \;\; \mathbf{V}^{t,l} = \mathbf{C}^{t,l}\mathbf{W}_v^{l},
\end{equation}
where each attention layer has its own projection matrices indexed by $l$. 
The hidden state $\mathbf{Z}$ is naturally a function of both the diffusion timestep and layer.
In a cross-attention layer, the text conditioning $\mathbf{C}=\mathbf{T}$ is not a function of $t$ and $l$ by default, but recent advances in textual inversion like NeTI~\cite{alaluf2023neural} show that this spacetime conditioning can improve personalization.

\paragraph{Mixture-of-Expert (MoE) Layer}
A MoE layer~\cite{shazeer2017,fedus2022switch} consists of $N$ expert networks and a router network that softly combines the output of the different experts:
\begin{gather}
    \mathbf{Z} = \sum_{n=1}^N \mathbf{R}_n \odot \text{Expert}_n(\mathbf{Z}), \;\; \\
    \mathbf{R} = \text{Router}(\mathbf{Z}) = \text{Softmax}(f(\mathbf{Z})), 
\end{gather}
where $\odot$ denotes the Hadamard product, and $\mathbf{R} \in \mathbb{R}^{l \times N}$. The router is a learned network that outputs a soft attention map over the input dimensions (i.e., latent pixels). Functionally, the router maps each latent pixel to $N$ logits that are then passed through a softmax. The mapping function $f$ can be a simple linear layer or \ka{an} MLP.

\subsection{Mixture-of-Attention Layer}
Under the general framework of MoE layers, our proposed MoA layer has two distinct features.  First, each of our `experts' is an attention layer, i.e. the attention mechanism and the learnable project layers described in~\cref{eqn:attn}.  
Second, we have only two experts, a frozen prior expert and a learnable personalization expert.  
Together, our MoA layer has the following form:
\begin{gather}
\mathbf{Z}^{t,l} = \sum_{n=1}^2 \mathbf{R}_n^{t,l} \odot \text{Attention}(\mathbf{Q}_n^{t,l},\mathbf{K}_n^{t,l},\mathbf{V}_n^{t,l}), \\
\mathbf{R}^{t,l} = \text{Router}^l(\mathbf{Z}^{t,l}).
\label{eqn:moa}
\end{gather}
Note, each MoA layer has its own router network, hence it is indexed by the layer $l$ and each attention expert has its own projection layers, hence they are indexed by $n$. 
We initialize both of the experts in a MoA layer using the attention layer from a pretrained model. The prior expert is kept frozen while the personalization expert is finetuned.

\begin{figure}[t]
    \centering
        \centering
        \includegraphics[width=.8\columnwidth]{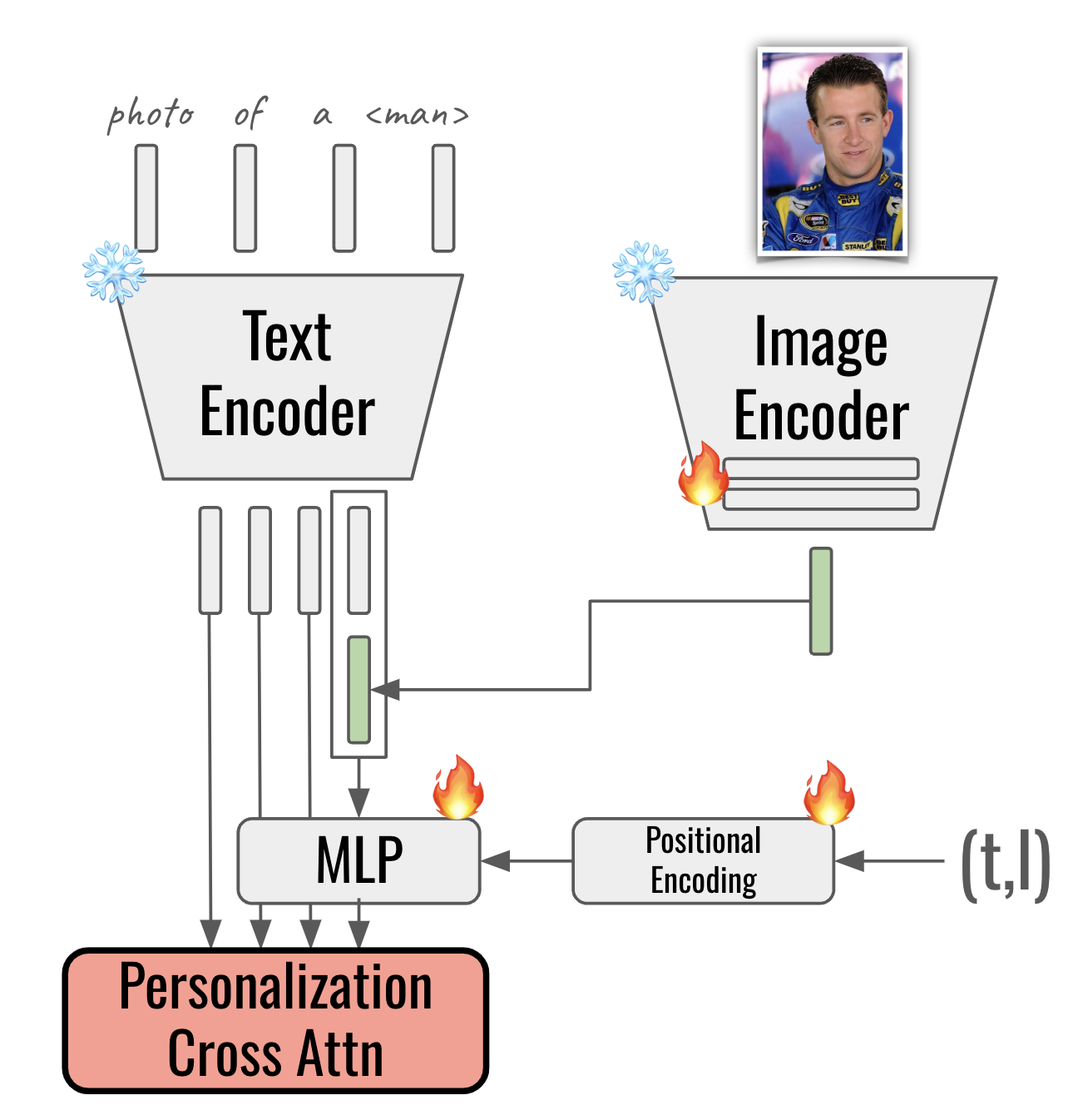}
        \caption{\textbf{Multimodal prompts.} Our architecture enables us to inject images as visual tokens that are part of the text prompt, where each visual token is attached to a text encoding of a specific token.}
        \label{fig:image-inject}
\end{figure}    

\begin{figure}
  \centering
  \includegraphics[width=\columnwidth]{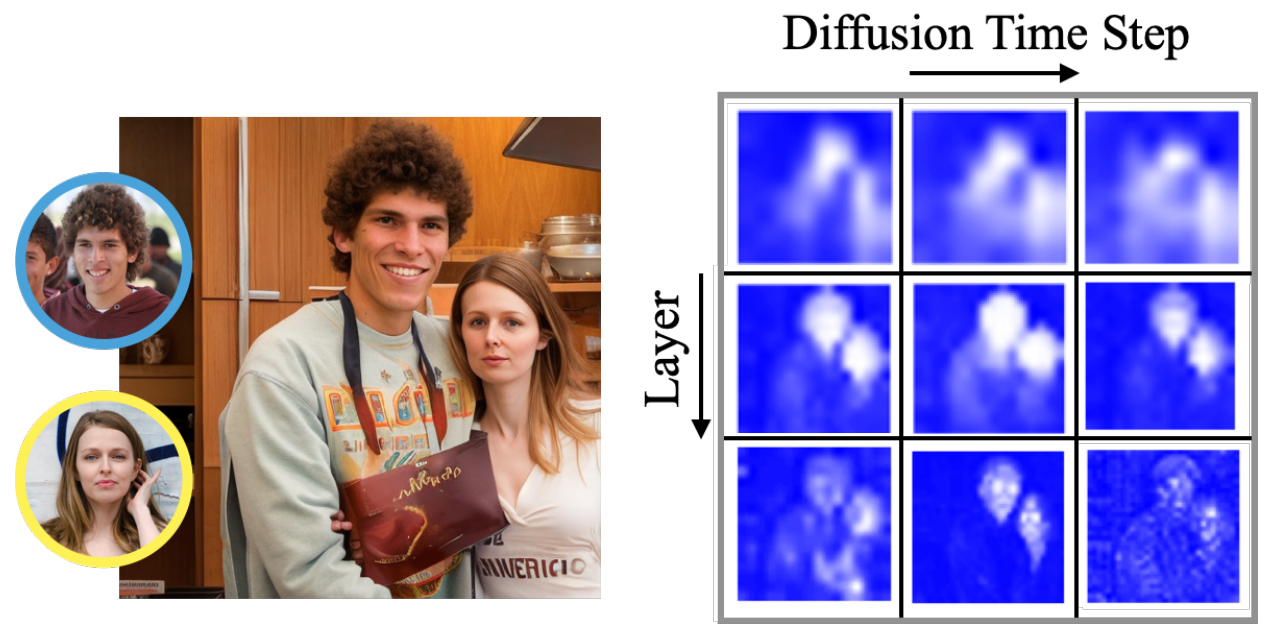}
  \caption{ \textbf{Router Visualization.} Our router learns to generate soft segmentation maps per time step in the diffusion process and per layer. Distinct parts of the subjects, in different resolutions, are highlighted across various time steps and layers.
  }
  \label{fig:router-viz}
\end{figure}

\subsubsection{Cross-Attention Experts}
While in the MoA self attention layers, the two experts receive the same inputs, in the MoA cross attention layers, the two experts take different inputs. To fully preserve the prior, the prior expert receives the standard text-only condition.  
To handle image inputs, the personalization expert receives a multi-modal prompt embedding described in the following section.  

\paragraph{Multimodal Prompts}
Given a subject image, $I$, it is injected into the text prompt as shown in~\cref{fig:image-inject}.  First, image feature, $\mathbf{f}$, is extracted using a pretrained image encoder (e.g. CLIP image encoder), $\mathbf{f} = E_\text{image}(I)$.
The image feature is concatenated with the text embedding at the corresponding token, $\mathbf{t}$, say the embedding of the `man' token. We refer to the concatenated embedding as the multi-modal embedding, $\mathbf{m} = \text{Concat}(\mathbf{f}, \mathbf{t})$.  
We further condition the multi-modal embedding on two information: the diffusion timestep, $t$, and U-Net layer $l$ through a learned positional encoding ($\text{PE}:\mathbb{R}^2 \mapsto \mathbb{R}^{2d_t}$) as follows
\begin{gather}
    \bar{\mathbf{m}}_{t,l} = \text{LayerNorm}(\mathbf{m}) + \text{LayerNorm}(\text{PE}(t, l)).
    \label{eqn:spacetime}
\end{gather}
From previous work on optimization-based personalization, the diffusion time and space conditioning can improve identity preservation~\cite{voynov2023p+,alaluf2023neural,zhang2023prospect}. Finally, the embedding is passed to a learnable MLP.

\begin{figure*}[t]
  \centering
  \includegraphics[width=\textwidth, trim=0 400 450 0, clip]{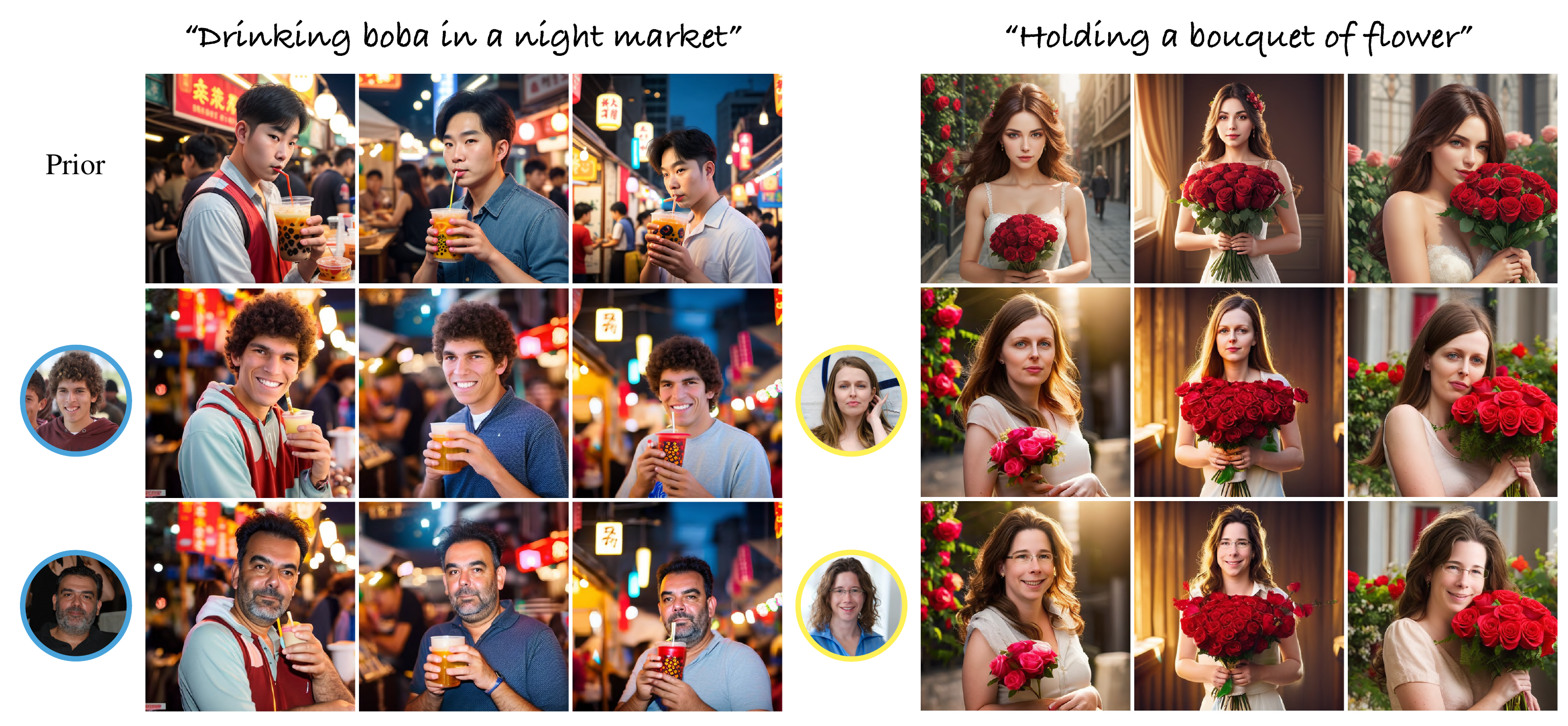}
  
  \caption{ \textbf{Disentangled subject-context control with a single subject.}  The top row is generetad using only the prior branch. Each column is a different random seed.  MoA allows for disentangled subject-context control. Namely, injecting different subjects lead to only localized changes to the pixels pertaining to the foreground human. 
  }
  \label{fig:qualitative-ss}
\end{figure*}

\subsection{Training}

\subsubsection{Training the Router}
The router is trained with an objective that encourages the background pixels (i.e. not belonging to the image subject) to utilize the ``prior'' branch.  The foreground pixels are not explicitly optimized w.r.t. any target.  The loss is computed after accumulating the router predictions at all layers:
\begin{gather}
    \mathcal{L}_\text{router} = \| (1-\mathbf{M}) \odot (1 - \mathbf{R}) \|^2_2 , \\
    \mathbf{R} = \frac{1}{|\mathbb{L}|} \sum_{l \in \mathbb{L}} \mathbf{R}_0^l,
\end{gather}
where $\mathbf{R}_0^l$ is the router weight for the prior branch at U-Net layer $l$, and $\mathbf{M}$ is the foreground mask of the subject. $\mathbb{L}$ is the set of layers we penalize in the U-Net, and $|\mathbb{L}|$ is the number of layers. 
In practice, we exclude the first and last block of the U-Net (i.e. first two and last three attention layers). Empirically, they encode low-level features of the image and are less relevant to the notion of subject versus context.  
Across different U-Net layers and diffusion timesteps, the personalization expert focuses on regions associated with the subject, and the prior branch accounts for most of the background while still having a base level of contribution to the subjects.  The routers also behave differently at different layers and different timesteps.  For example, the personalization expert at one layer/timestep might attends to the face while at another layer/timestep attneds to the body, as visualized in \cref{fig:router-viz}.
.

\subsubsection{Overall Training Scheme}
Typically, training or finetuning diffusion models is done by using the full (latent) image reconstruction loss, $\mathcal{L}_\text{full}(\mathbf{Z}, \hat{\mathbf{Z}}) = \|\mathbf{Z} - \hat{\mathbf{Z}}\|_2^2$.  
In contrast, recent personalization methods use the segmentation mask for masked (foreground only) reconstruction loss to prevent the model from confounding with the background, i.e. $\mathcal{L}_\text{masked}(\mathbf{Z}, \hat{\mathbf{Z}}) = \|\ \mathbf{M} \odot (\mathbf{Z} - \hat{\mathbf{Z}})\|_2^2$. 
In previous training-based methods like Fastcomposer~\cite{xiao2023fastcomposer}, they need to balance between preserving the prior by using the full image loss with focusing on the subject by using the masked loss, i.e. $\mathcal{L} = p\mathcal{L}_\text{full} + (1-p)\mathcal{L}_\text{masked} $ where $p$ is the probability of using the full loss and was set to $0.5$.
Because of our MoA layer, we do not need to trade-off between prior and personalization. 
Hence, we can use the best practice for the personalization, which is only optimizing the foreground reconstruction loss.  Our frozen prior branch naturally plays the role of preserving the prior. 
Our finetuning consists of the masked reconstruction loss, router loss, and cross attention mask loss:
\begin{gather}
    \mathcal{L} = \mathcal{L}_\text{masked} + \lambda_r \mathcal{L}_\text{router} + \lambda_o \mathcal{L}_\text{object},
\end{gather}
\kcwqa{where the $\mathcal{L}_\text{object}$ is the balanced L1 loss proposed in Fastcomposer~\cite{xiao2023fastcomposer}.  We apply it to our personalization experts: 
\begin{gather}
    \mathcal{L}_\text{object} =\frac{1}{|\mathbb{L}||\mathbb{S}|} \sum_{l \in \mathbb{L}}\sum_{s \in \mathbb{S}} \text{mean}( (1-\mathbf{M}_s) \odot \big(1 - \mathbf{A}^l_s)\big) - \text{mean}\big(\mathbf{M}_s \odot \mathbf{A}^l_s\big)  , 
\end{gather}
where $\mathbb{S}$ denotes the set of subjects in an image, $\mathbf{M}_s$ the segmentation mask of subject $s$, and $\mathbf{A}^l_s$ the cross attention map at the token where subject $s$ is injected.
}

\section{Experiments}

In this section, we show results to highlight MoA's capability to perform disentangled subject-context control, handle occlusions through both qualitative and quantitative evaluations.  We also show analysis of the router behavior as a explanation for the new capability.
\subsection{Experimental Setup}
\paragraph{Datasets.} For training and quantitative evaluation, two datasets were used.  For training, we used the FFHQ~\cite{karras2019style} dataset preprocessed by~\cite{xiao2023fastcomposer}, which also contains captions generated by BLIP-2~\cite{li2023blip} and segmentation masks generated by MaskedFormer~\cite{cheng2022masked}. For the test set of the quantitative evaluation, we used test subjects from the FFHQ dataset for qualitative results, and 15 subjects from the CelebA dataset~\cite{liu2015deep} for both of the qualitative and quantitative evaluation following previous works. 

\paragraph{Model details.}
For the pretrained T2I models, we use StableDiffusion v1.5 ~\cite{rombach2022high}. For some qualitative results, we use the community finetuned checkpoints like \texttt{AbsoluteReality\_v1.8.1}.
For the image encoder, we follow previous studies and use OpenAI's \texttt{clip-vit-large-patch14} vision model. 
We train our models on 4 NVIDIA H100 GPUs, with a constant learning rate of 1e-5 and a batch size of 128. 
Following standard training for classifier-free guidance~\cite{ho2022classifier}, 
we train the model without any conditions 10\% of the time.
During inference, we use the UniPC sampler~\cite{zhao2024unipc}.

\begin{figure*}[t]
  \centering

  \includegraphics[width=\textwidth, trim=0 350 380 0, clip]{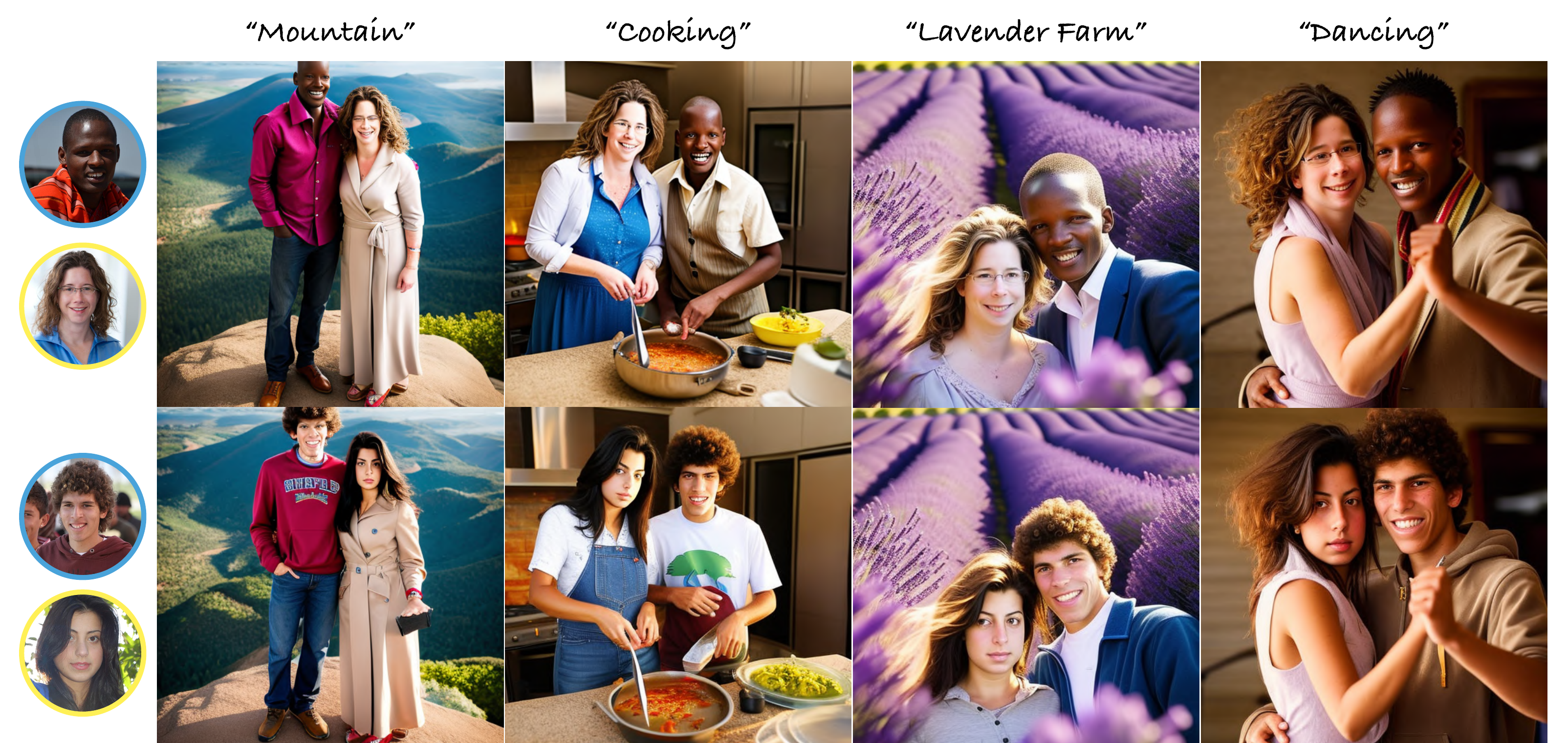}
  \caption{ \textbf{Images with close interactions of two subjects.} MoA can generate images with different subject layouts and different interaction types among the subjects.}
  \label{fig:qualitative-couple-interaction}
\end{figure*}

\subsection{Results}

All the results presented in this section are performed on the held-out test subjects of the FFHQ dataset. 

\paragraph{Disentangled subject-context control.}
The first major result is the disentangled subject-context control that is enabled by our MoA architecture.  
We show unforeseen disentanglement between the background control using the random seed and input image subjects all in the single forward pass. 
In~\cref{fig:qualitative-ss}, we show the results of someone drinking boba in a night market and holding a bouquet of roses. 
Notice that as we change the input subjects while holding the seed constant, we are able to perform localized subject change without affecting the background. 
Moreover, in the top row, we show samples from using only the prior branch. 
The content is preserved after we inject different subjects. 
This allows for a new application where the users can quickly generate images and select them by content, and then inject the personalized information. 
They can then also easily swap the input subjects.

\paragraph{Image quality, variation, and consistency.}
Another unique aspect about MoA is the ``localized injection in the prompt space''.  
This feature allows for a surprising ability to handle occlusion. 
In~\cref{fig:qualitative-ss}, we can see the boba and roses occluding part of the face and body.  Despite the occlusion, the face details are preserved, and the body is also consistent with the face. For example, the men holding the boba have a consistent skin tone and texture in the arms as suggested by their face. 
We show additional results of handling occlusion in~\cref{fig:qualitative-ss-portrait} where we generate portraits photos with different costumes. 
In the generated images, a large portion of the face can be occluded.  Our method is able to handle such cases while preserving the identity.

\begin{figure*}[t]
  \centering
\includegraphics[width=\textwidth,trim=0 500 130 0, clip]{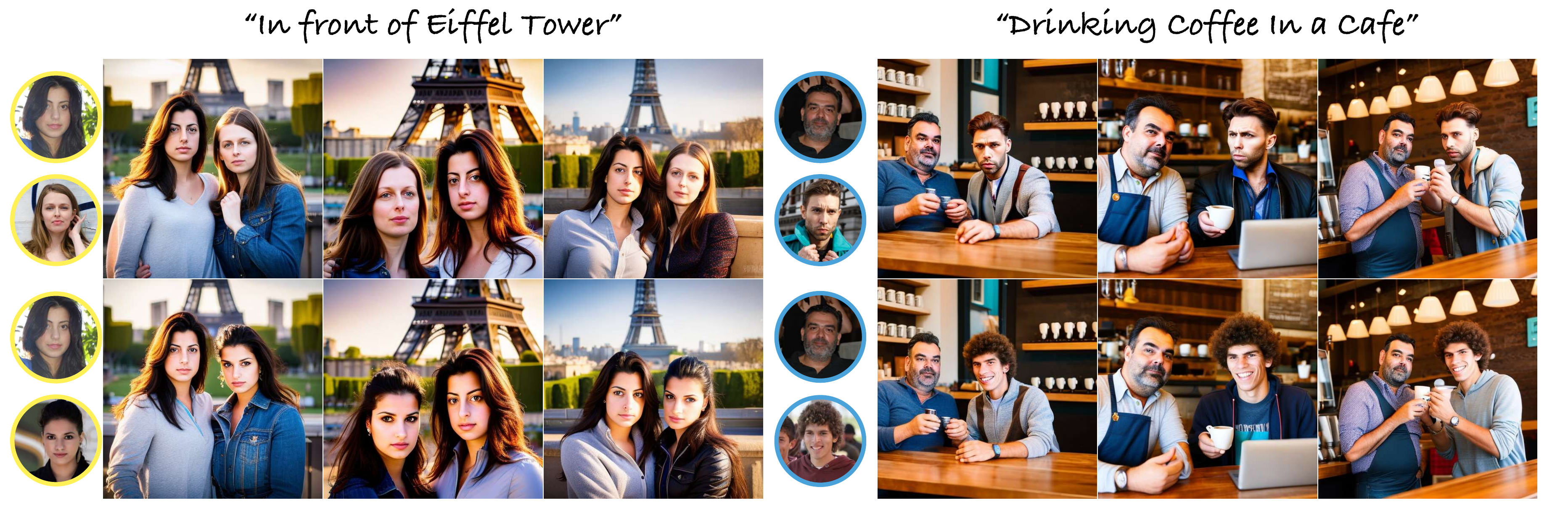}
  \caption{ \textbf{Disentangled subject-context control with a multiple subjects.} MoA retains the disentangled subject-context control in the multi-subject scenario.  One or both of the subjects can be swapped without substantial effect on the context.}
  \label{fig:qualitative-couple-disentangled}
\end{figure*}

\begin{figure}[t]
  \centering
  \includegraphics[width=\columnwidth,trim=0 500 800 0, clip]{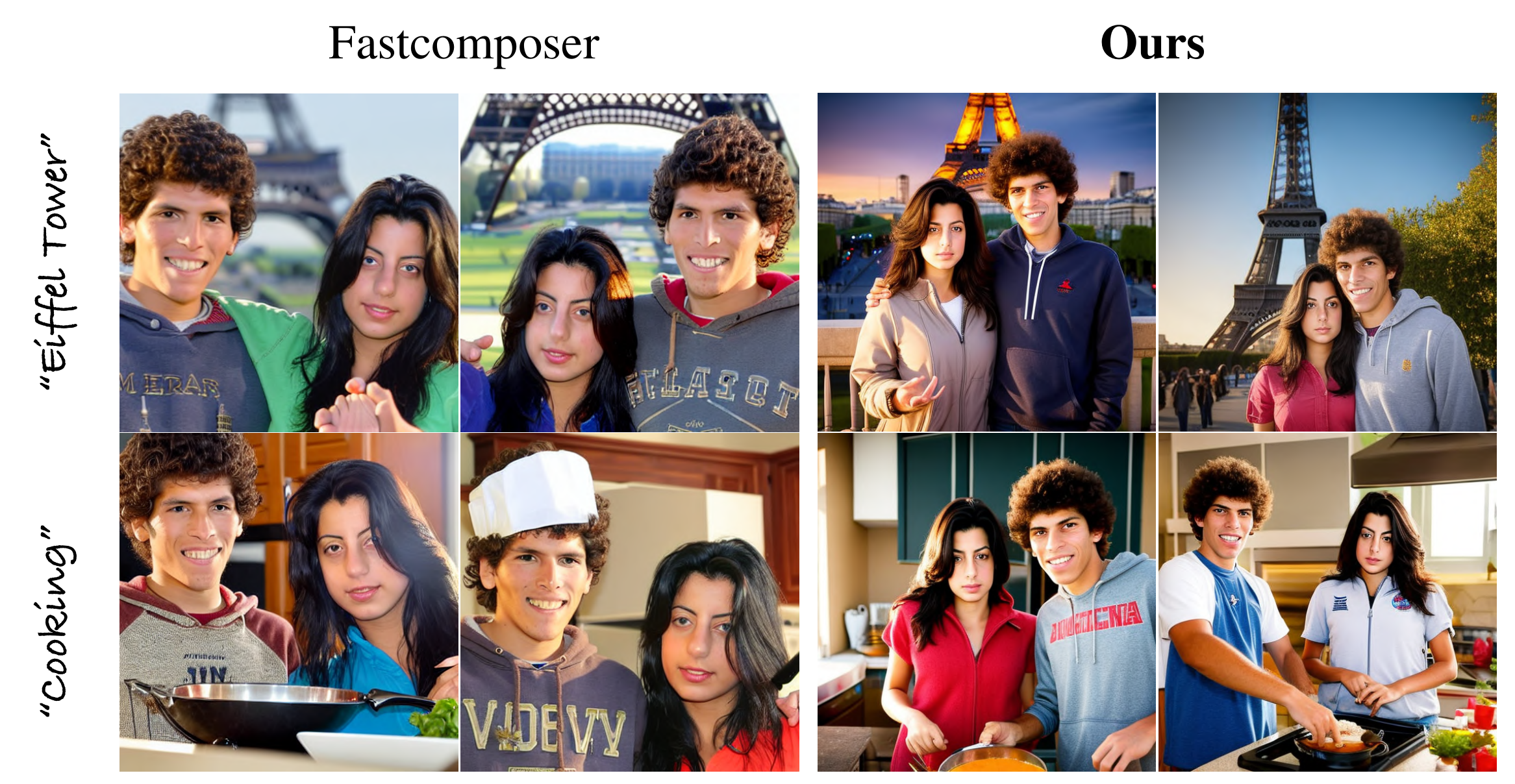}
  \caption{ \textbf{Comparison with Fastcomposer in the multi-subject setting.} }
  \label{fig:multi-subject-comparison}
\end{figure}

\paragraph{Multi-subject composition.}
This ability to generate full-body consistent subjects and handle occlusion unlocks the ability generate multi-subject images with close interaction between subjects.
In ~\cref{fig:qualitative-couple-interaction}, we show generated photos of couples with various prompts.  Even in cases of dancing, where the subjects have substantial occlusion with each other, the generation is still globally consistent (i.e. the skin tone of the men's arms match their faces).  
Furthermore, in~\cref{fig:qualitative-couple-disentangled}, we show that the \emph{disentangled subject-context control} capability still holds in the multi-subject case. This allows the users to swap one or both of the individuals in the generated images while preserving the interaction, background, and style. 
Lastly, when comparing our results with Fastcomposer in the multi-subject setting, MoA is able to better preserve the context and produce more coherent images (see~\cref{fig:multi-subject-comparison}). While Fastcomposer is able to inject multiple subjects and modify the background, the subjects are not well integrated with the context. This is evident in cases where the prompt describes an activity, such as ``cooking''.

\paragraph{Analysis.}

For analysis, we visualize the router prediction in~\cref{fig:router-viz}.  
We visualize the behavior using the same random seed, but different input subjects. 
The behavior of the router is consistent across the two subject pairs, and route most of the background pixel to the prior branch. 
We believe this explains why MoA allows for disentangled subject-context control.
See the supplementary material for more visualization with different random seeds where the router behavior changes, hence leading to a different layout and background content.

\section{Applications}
\label{sec:app}

In this section, we demonstrate a number of applications enabled by the disentangled control of MoA and its compatibility with existing image generation/editing techniques developed for diffusion-based models.   
In particular, the simplicity of the design in MoA makes it compatible with using ControlNet (\cref{sec:app:controlnet}).  
MoA can create new characters by interpolating between the image features of different subjects, which we refer to as subject morphing (\cref{sec:app:subject_morph}).
Beyond generation, MoA is also compatible with real-image editing techniques based on diffusion inversion~\cite{song2020denoising,mokady2023null,Dhariwal2021DiffusionMB} (\cref{sec:app:real_image_edit}).
We include three more applications (style swap with LoRA, time lapse, and consistent character storytelling) in ~\cref{sec:app:app}.

\subsection{Controllable Personalized Generation}
\label{sec:app:controlnet}
\begin{figure*}[t]
  \centering
  \includegraphics[width=\textwidth, trim=0 400 300 0, clip]{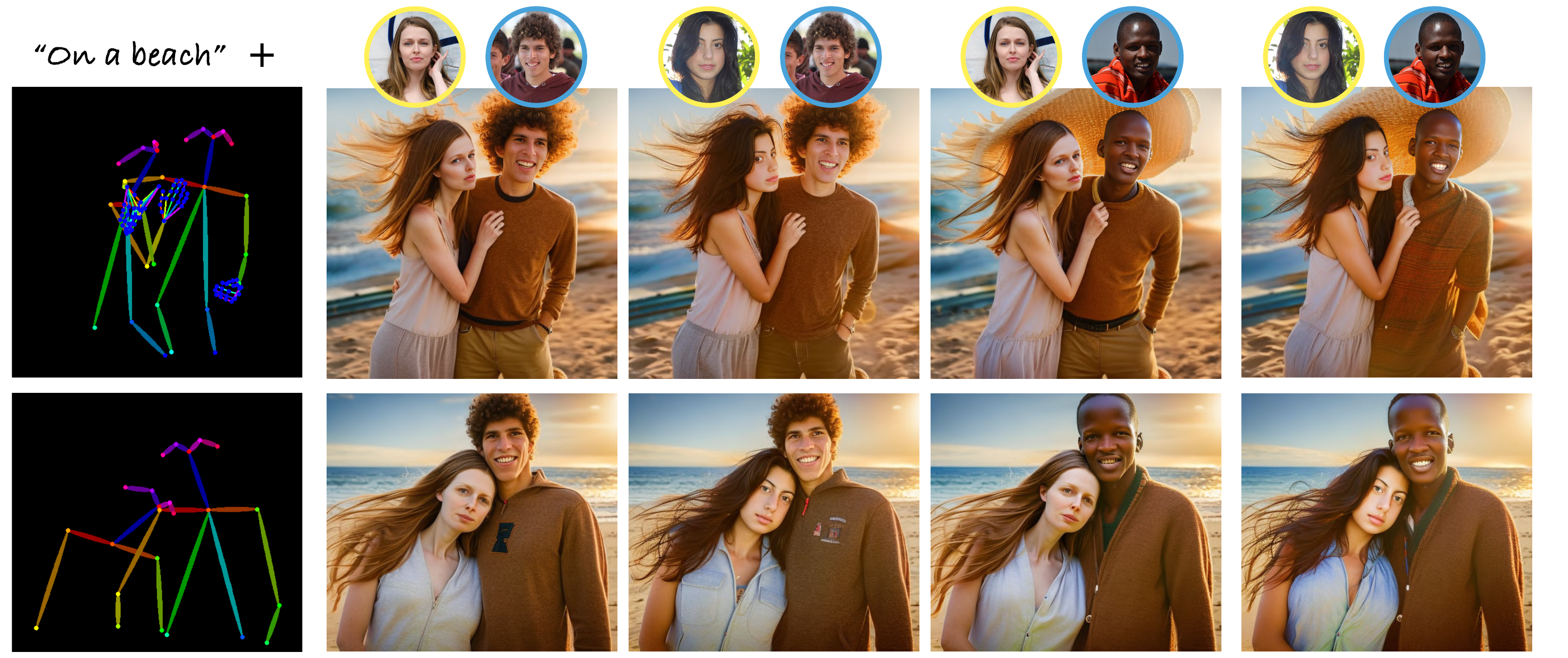}
  \caption{ \textbf{Controllable personalized generation.}  MoA is compatible with ControlNet.  Given the same prompt, the user can use ControlNet for pose control.  In this application, MoA still retains the disentangled subject-context control.  
  }
  \label{fig:controlnet}
\end{figure*}

A key feature of MoA is its simplicity and minimal modification to the base diffusion model.  This makes it naturally compatible with existing extensions like ControlNet~\cite{zhang2023adding}.  Since MoA operates only within the attention mechanism, the semantics of the latent are preserved in-between U-Net blocks, where ControlNet conditioning is applied. This makes it possible to use ControlNet in exactly the same way as it would be used with the prior branch of the model. In~\cref{fig:controlnet}, we show examples of adding pose control to MoA using ControlNet.  Given the same text prompt and random seed, which specifies the context, the user can use ControlNet to change the pose of the subjects.  Even in such a use case, MoA retains the disentangled subject-context control and is able to swap the subjects.

\subsection{Subject Morphing}
\label{sec:app:subject_morph}

\begin{figure*}[t]
  \centering
  \includegraphics[width=\textwidth]{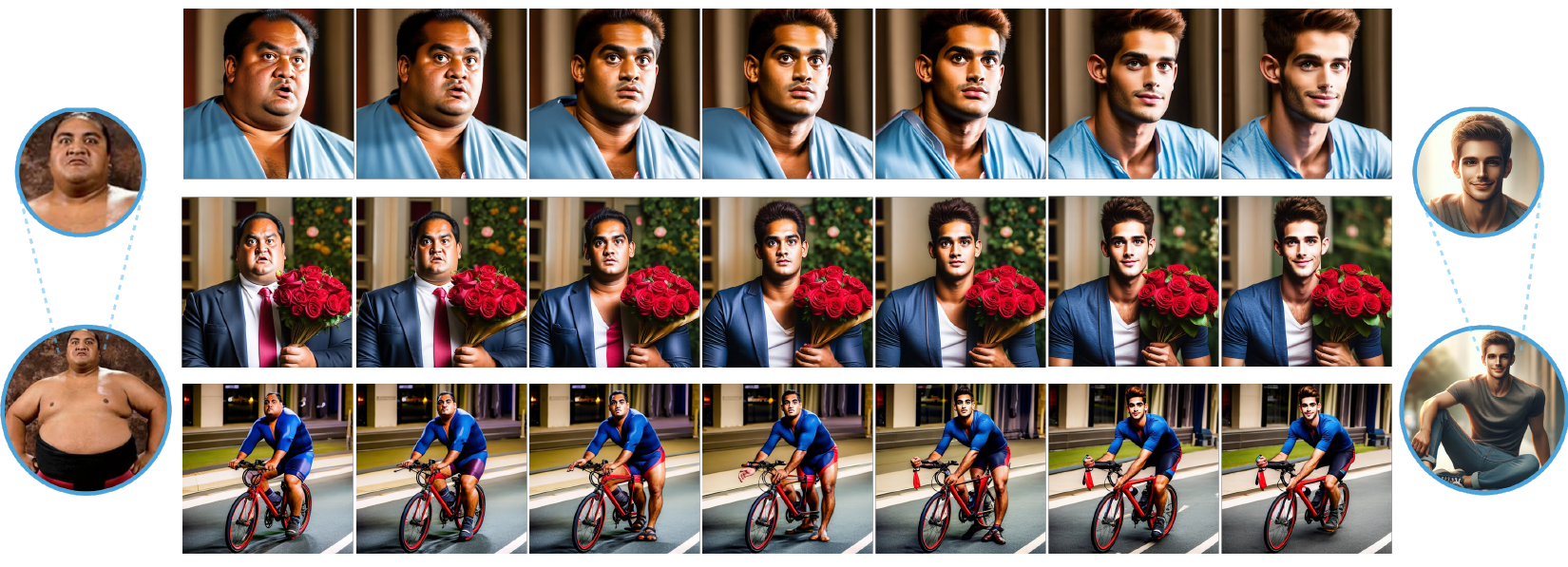}    
  \caption{ \textbf{Subject morphing.} By interpolating between the embeddings of the image encoder in MoA, we can achieve a morphing effect between two subjects with different characteristics. On the left is the image 'Yokozuna', and on the right is an image generated by DALL-E 3.
  }
  \label{fig:subject_morph}
\end{figure*}

By interpolating the image feature outputted by the learned image encoder in MoA, one can interpolate between two different subjects.  Since MoA encodes more than the face of the subject and has a holistic understanding of the body shape and skin tone, we are able to interpolate between two very different subjects.  In~\cref{fig:subject_morph}, we interpolate between the Yokozuna, who has a large body and darker skin tone, and a generated male character, who has a smaller body and a pale skin tone.  The features of the interpolated subjects are preserved with different prompts like `holding a bouquet' and `riding a bike'.

\subsection{Real Image Subject Swap}
\label{sec:app:real_image_edit}



\begin{figure}[t]
  \centering

  \includegraphics[width=\columnwidth, trim=0 0 1200 0, clip]{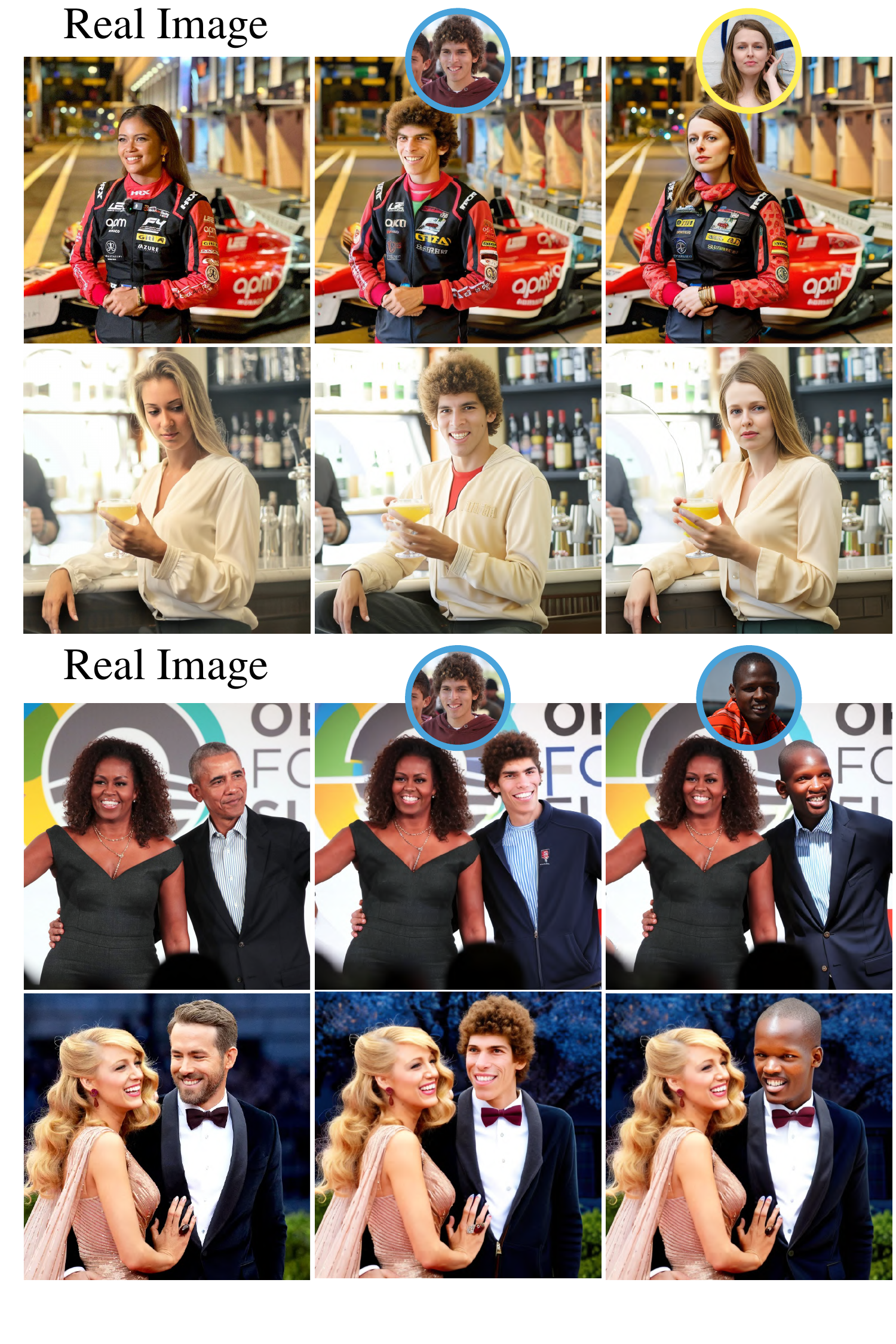}    
  \caption{ \textbf{Real image editing with MoA.} MoA is compatible with diffusion-based image editing techniques with DDIM Inversion.  Starting with the inverted noised, MoA is able to replace the subject in the reference image.
  }
  \label{fig:real_image_editing}
\end{figure}

Thanks to the simplicity of MoA, and the minimal deviation from the prior model, it can be used in conjunction with DDIM Inversion~\cite{song2020denoising,mokady2023null} to enable real-image editing.  \cref{fig:real_image_editing} shows results of this application.  
In the case of a single subject photo, we run DDIM Inversion with the prompt ``a person''.  
Starting from the inverted random noise, we run generation using MoA and inject the desired subject in the `person' token.  
For swapping a subject in a couple photo, we run DDIM inversion with the prompt ``a person and a person''.  During MoA generation, we used a crop of the subject to keep in the first `person' token, and inject the desired subject image into the second `person' token.

\section{Limitations}

\begin{figure}[t]
    \centering
    \includegraphics[width=\columnwidth,trim=0 0 850 0, clip]{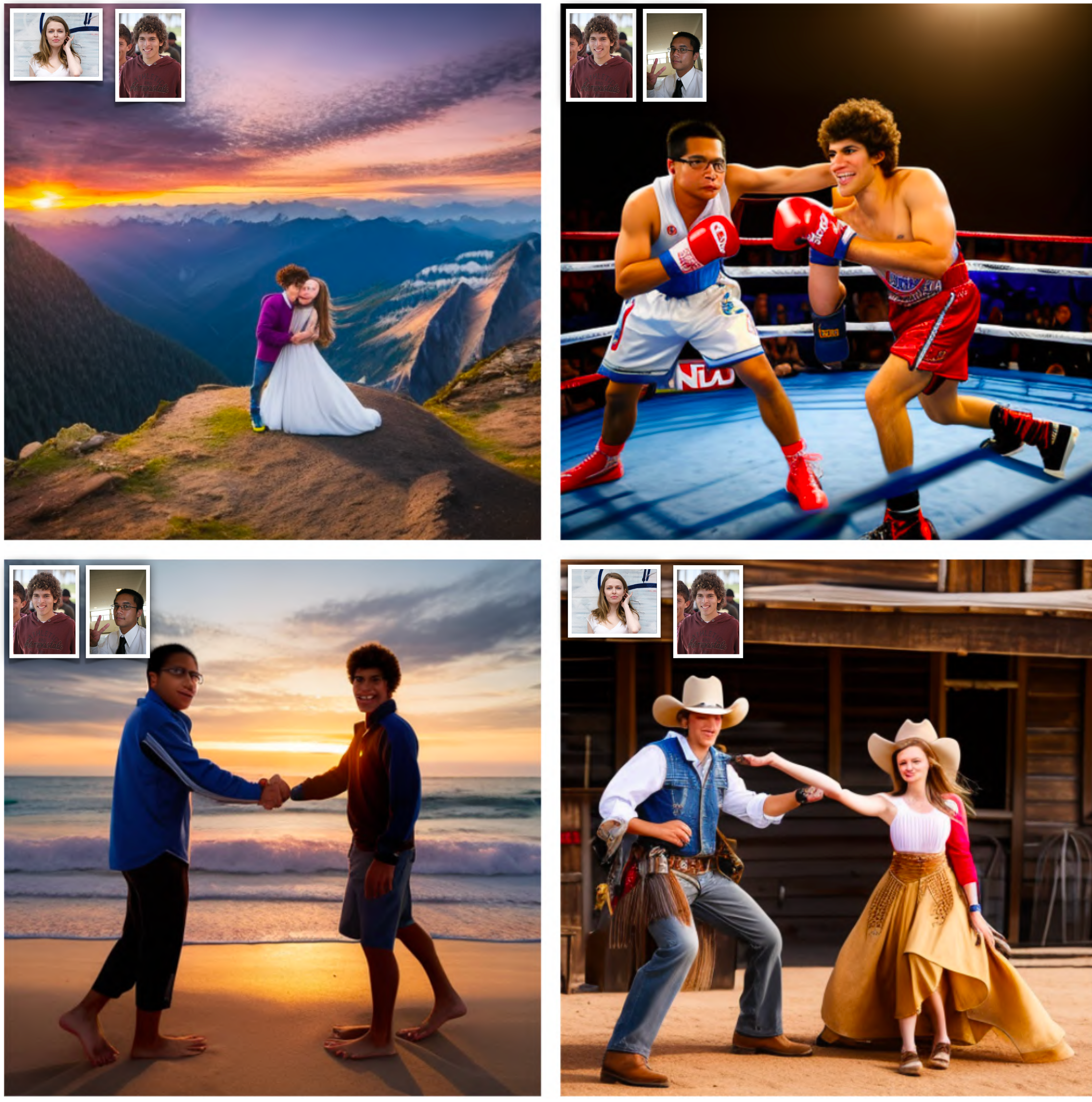}
      \captionof{figure}{\textbf{Limitation.} One key feature of MoA is enabling the generation of images with complex interaction scenarios, which result in full-body images.  They inevitably contain small faces, which remains a hard task for the underlying Stable Diffusion model.}
    \label{fig:limitation}
\end{figure}

Firstly, due to inherent limitations of the underlying Stable Diffusion model, our method sometimes struggles with producing high-quality small faces (see~\cref{fig:limitation}). This particularly affects the ability to depict multiple people in the same image, as scenes involving interactions typically demand a full-body photo from a longer distance, rather than an upper-body portrait. Secondly, generating images that depict intricate scenarios with a wide range of interactions and many individuals continues to be a challenging task. This difficulty is again largely due to the inherent limitations of Stable Diffusion and CLIP, particularly their inadequate grasp of complex compositional concepts, such as counting objects.
\kcwqa{
Specifically for MoA, the current implementation has limited ability to perform text-based expression control.  Since during finetuning, the expression in the input subject image and the output reconstruction target are the same, the model entangled the notion of `identity' and `expression'.  A future direction worth exploring is to use a slightly different input image; for example, two different frames from a video explored in other topics~\cite{kulal2023putting}.
}

\section{Conclusion}

We introduce Mixture-of-Attention (MoA), a new architecture for personalized generation that augments a foundation text-to-image model with the ability to inject subject images while preserves the prior capability of the model. 
While images generated by existing subject-driven generation methods often lack diversity and subject-context interaction when compared to the images generated by the prior text-to-image model, MoA seamlessly unifies the two paradigms by having two distinct experts and a router to dynamically merges the two pathways. 
MoA layers enables the generation of personalized context from multiple input subjects with rich interactions, akin to the original non-personalized model, within a single reverse diffusion pass and without requiring test-time fine-tuning steps, unlocking previously unattainable results. 
In addition, our model demonstrates previously unseen layout variation in the generated images and the capability to handle occlusion from objects or other subjects, and handle different body shapes all without explicit control.
Lastly, thanks to its simplicity, MoA is naturally compatible with well-known diffusion-based generation and editing techniques like ControlNet and DDIM Inversion.  As an example, the combination of MoA and DDIM Inversion unlocks the application of subject swapping in a real photo.  
Looking ahead, we envision further enhancements to the MoA architecture through the specialization of different experts on distinct tasks or semantic labels. Additionally, the adoption of a minimal intervention approach to personalization can be extended to various foundational models (e.g. video, and 3D/4D generation), facilitating the creation of personalized content with existing and futuristic generative models.

\section*{Acknowledgement}

The authors would like to acknowledge Colin Eles for infrastructure support, Yuval Alaluf, Or Patashnik, Rinon Gal, Daniel Cohen-Or for their feedback on the paper, and other members on the Snap Creative Vision team for valuable feedback and discussion throughout the project.

\clearpage
\bibliographystyle{ACM-Reference-Format}
\bibliography{main}

\clearpage
\appendix

\section{Additional Experimental Details}
\paragraph{Finetuning hyperaparemters.} Training is done using the Accelerate library~\cite{accelerate} with 4 GPUs in mixed precision (\texttt{bf16}).
\cref{table:hyperp} summarizes the finetuning hyperparameters.  
\begin{table}[h]
\centering
\caption{Prompts used for generating the qualitative results.}
\label{table:hyperp}

\begin{tabular}{lc}
\toprule
\textbf{Name}                     & \textbf{Value} \\
\midrule
Training iteration                & 40k            \\
Batch size per GPU                & 32             \\
\# of GPUs                        & 4              \\
Learning rate                     & 5e-05       \\
Router regularization weight ($\lambda_r$)      & 1e-04       \\
Object regularization weight ($\lambda_o$)    & 1e-04       \\
Prob. of removing condition       & 0.1            \\
Prob. of using masked recon. loss & 1              \\
Max training diffusion timestep sampled    & 800           \\
\bottomrule
\end{tabular}
\end{table}

\paragraph{Prompts.} For generating the qualitative results, we use the prompts listed in~\cref{table:prompts}.  The special token `man' is replaced with `woman' when appropriate.

\begin{table}[h]
\centering
\small
\caption{Prompts used for generating the qualitative results.}
\label{table:prompts}
\begin{tabular}{r|p{6cm}}
\toprule
\textbf{ID}                     & \textbf{Prompt} \\
\midrule
vanGogh & :\texttt{a \textbf{man} and a \textbf{man} in the style of Van Gogh painting} \\
sofa & :\texttt{a \textbf{man} and a \textbf{man} sitting on a sofa} \\
book & :\texttt{a \textbf{man} and a \textbf{man} reading a book} \\
paris & :\texttt{a \textbf{man} and a \textbf{man} in Paris, shaking hands, with the Eiffel Tower in the background} \\
mountain & :\texttt{a \textbf{man} and a \textbf{man} standing on a mountain} \\
running & :\texttt{a \textbf{man} and a \textbf{man} running in a park} \\
lavender farm & :\texttt{a \textbf{man} and a \textbf{man} in a lavendar farm} \\
lab & :\texttt{a \textbf{man} in a laboratory} \\
bike & :\texttt{a \textbf{man} riding a bike} \\
cowboy & :\texttt{a portrait photo of a \textbf{man} dressed as a cowboy in the wild west, in from of a saloon, very high quality, professional photo, beautiful lighting, 8k} \\
spaceman & :\texttt{a close up portrait photo of a \textbf{man} dressed in a space suit without helmet, on Mars, in front of spacehip, beautiful starry sky, very high quality, professional photo, beautiful lighting, 8k} \\
racecar\_driver & :\texttt{a portrait photo of a \textbf{man} as a race car driver, very high quality, professional photo, beautiful lighting, 8k} \\
\bottomrule
\end{tabular}
\end{table}


\section{Ablation}
In this section, we ablate the model by (i) removing our primary contribution, the MoA layer, (ii) remove the community checkpoint and use the vanilla SD15 (i.e.\texttt{runwayml/stable-diffusion-v1-5}) checkpoint, and (iii) removing the image feature spacetime conditioning (\cref{eqn:spacetime}).  

\begin{figure}[h]
  \centering
  \includegraphics[width=\columnwidth, trim=430 50 350 0,  clip]{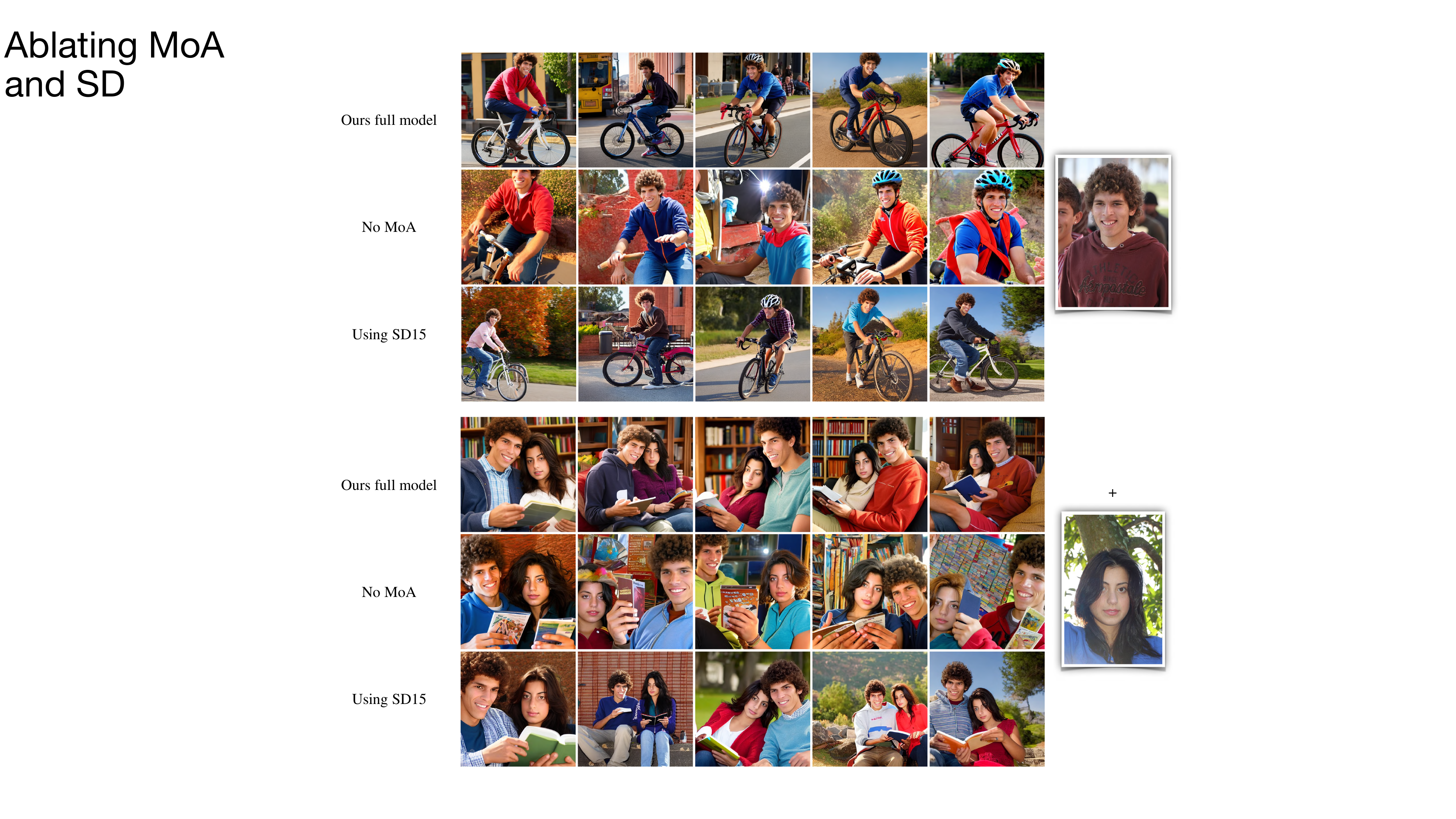}
  \caption{ Ablations: removing the MoA layer, and using the vanilla SD15 checkpoint. Constrasting with and without MoA layer, we clearly see that difference in context preservation.  Without the MoA layer, the context (i.e. object, background and interaction) is lost despite the foreground being preserved well.  Comparing our full model using the \texttt{AbsoluteReality} checkpoint with using the vanilla SD15 checkpoint, the behavior is similar, but overall texture differs. 
  }
  \label{fig:abla-moa-sd}
\end{figure}

In~\cref{fig:abla-moa-sd}, we can clearly see that removing the MoA layer produced images with substantially worse quality.  While the foreground subjects are well preserved, the context is mostly lost. 
When comparing to images generated using the base checkpoint, the behavior is similar (i.e. both subject and context are well preserved).  The primary difference is that the community checkpoint generates images with better overall texture.  Similar to recent works in subject-driven generation~\cite{yan2024perflow,po2023orthogonal,gu2023mix}, we use the community checkpoint because of their better texture. 

\begin{figure}[h]
  \centering
  \includegraphics[width=\columnwidth, trim=420 50 160 0,  clip]{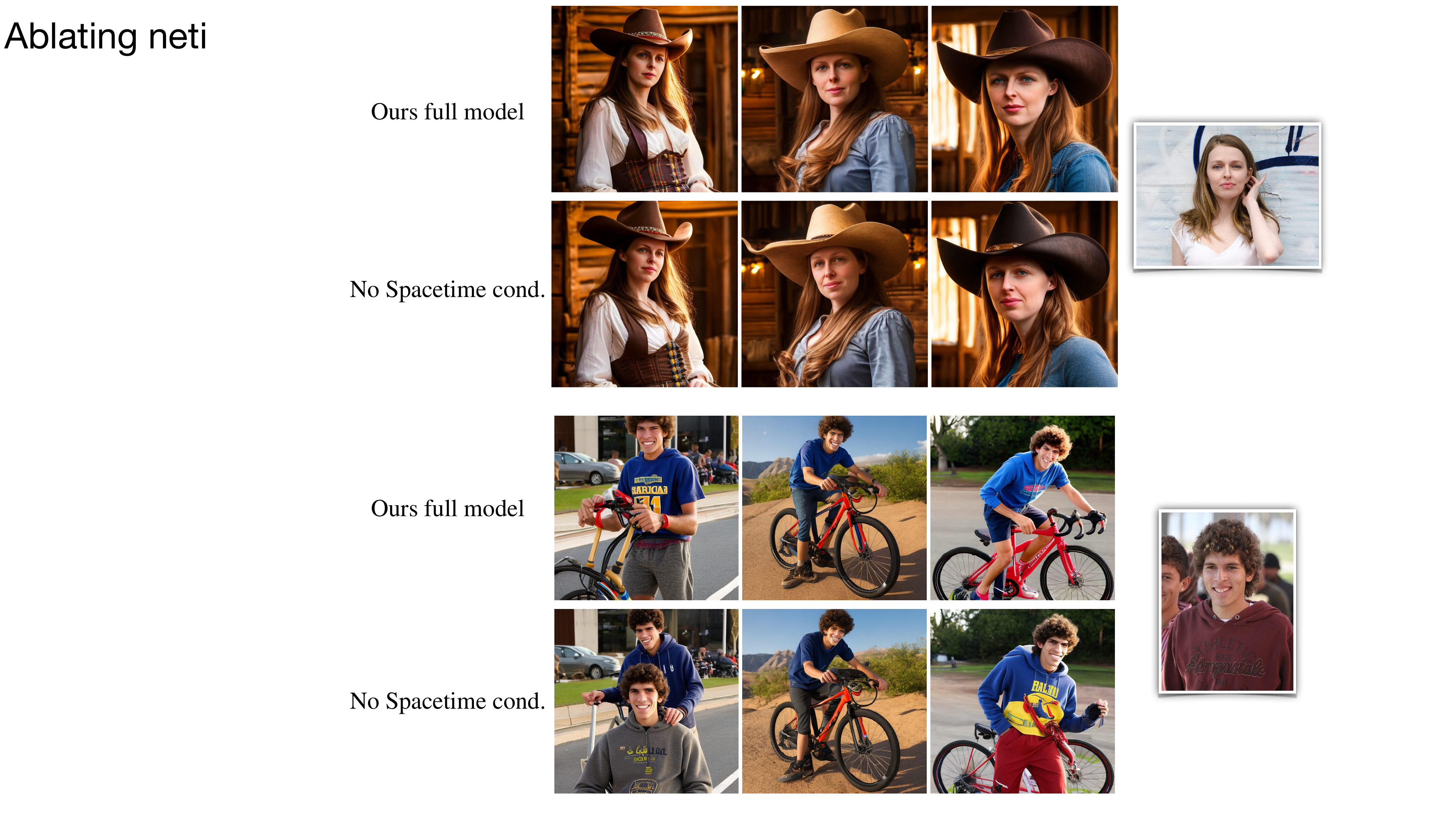}
  \caption{ 
  Ablations: removing the spacetime conditioning in the image feature.  Not having the spacetime conditing restricts the model, which results in worse identity preservation (top), and worse context preservation (bottom).
  }
  \label{fig:abla-spacetime}
\end{figure}

In~\cref{fig:abla-spacetime}, when we remove the spacetime conditioning (\cref{eqn:spacetime}) of the image features, this model is more restricted than our full model.  Intuitively, not having the spacetime conditioning makes the model worse at identity preservation.  This indirectly affects the overall model's capability at preserving the context as well.  

\section{Additional Qualitative Results}

\begin{figure}[t]
  \centering
  \includegraphics[width=\columnwidth, trim=0 100 450 0, clip]{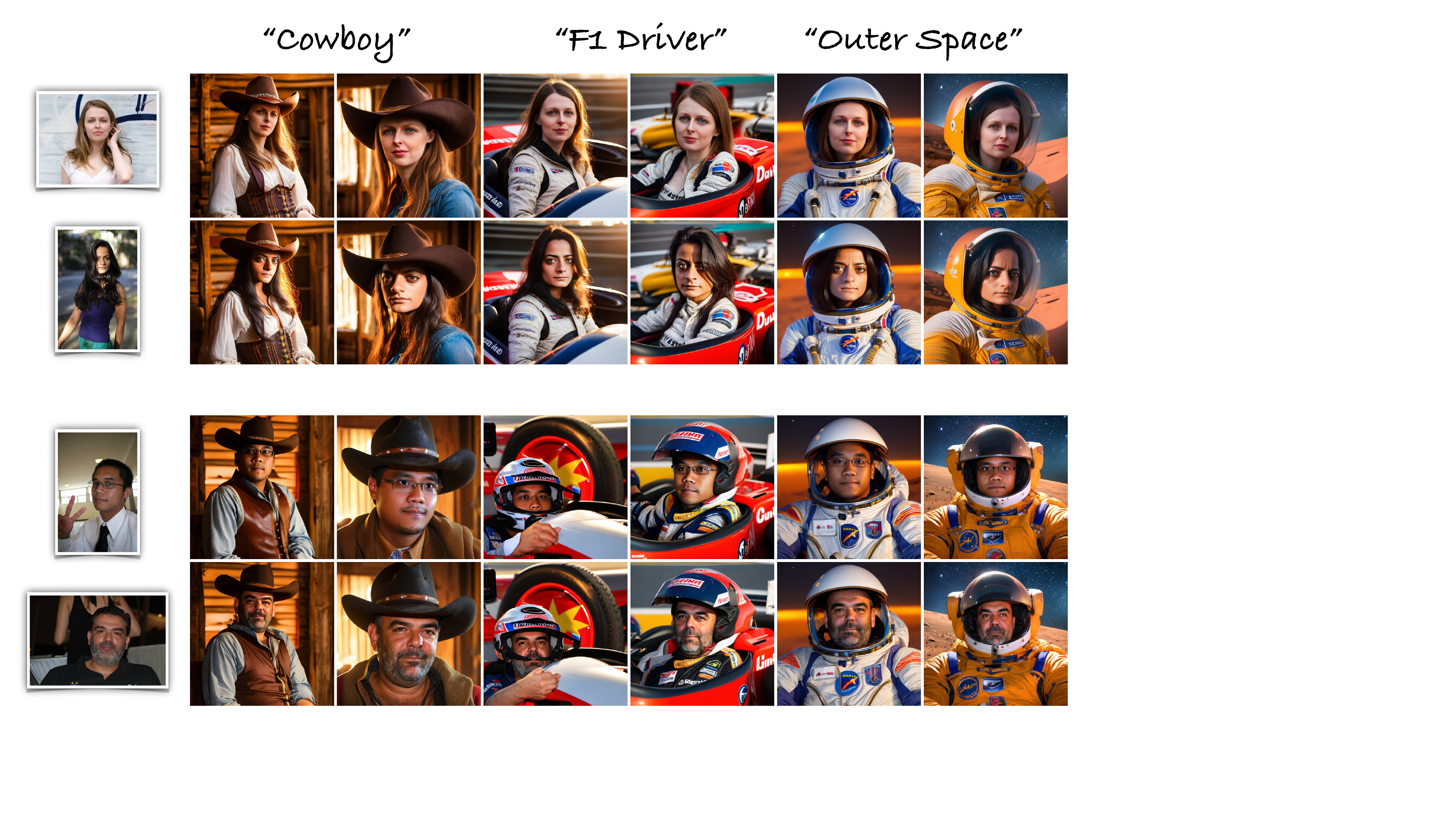}
  \caption{ \textbf{Single-subject portraits.}
  Our method is able to generate high-quality images of the input subjects in various imaginary scenarios and costumes.}
  \label{fig:qualitative-ss-portrait}
\end{figure}

\subsection{Handling Different Body Shapes}
\begin{figure}[h]
  \centering
  \includegraphics[width=\columnwidth, trim=0 50 230 0, clip]{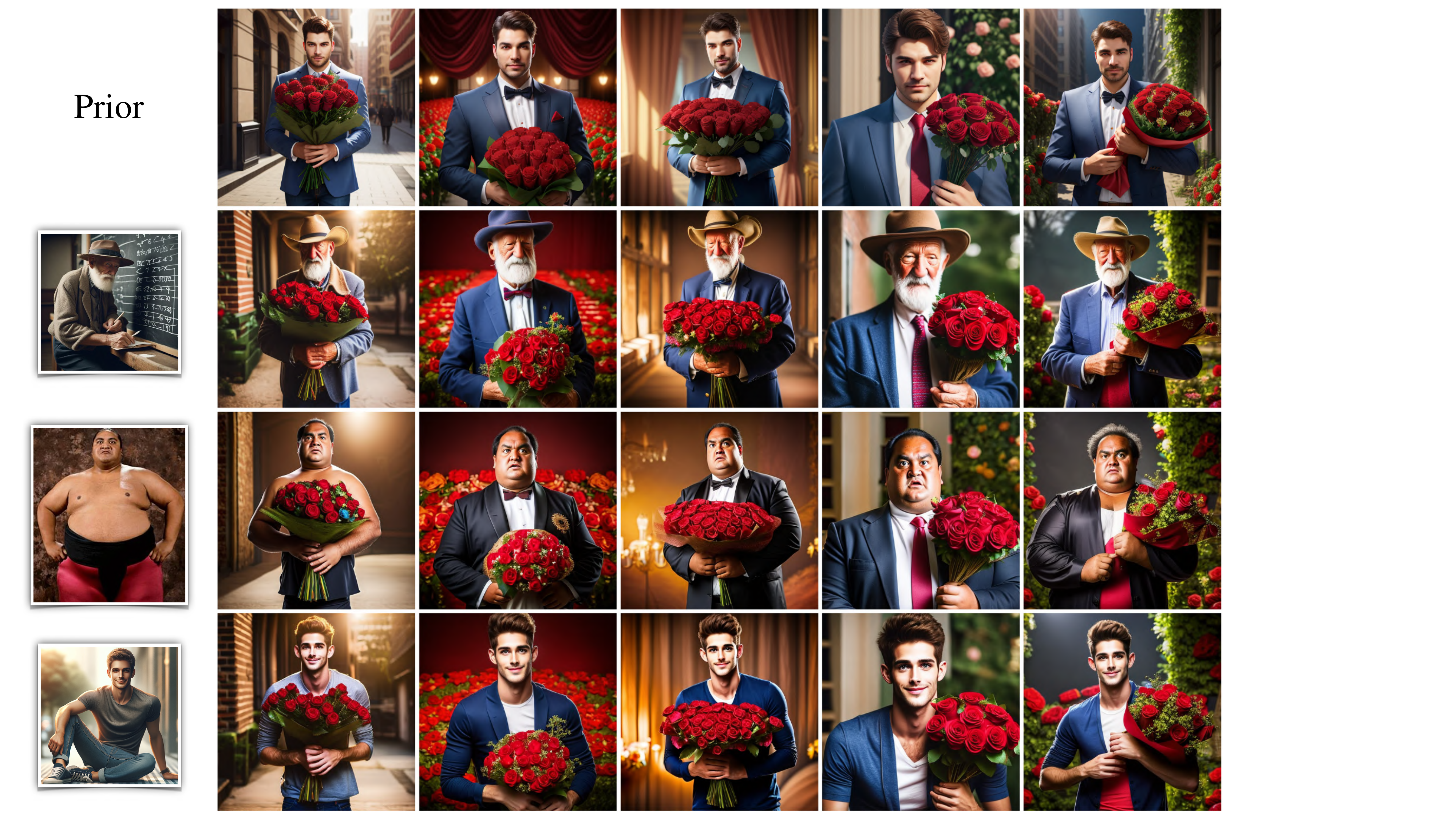}
  \caption{ Handling different body shapes.  
  }
  \label{fig:bodyshapes}
\end{figure}

Given MoA's capability to preserve both the subject and context well, we found a surprising use case where subjects with different body shapes can be injected, and the body shape is naturally preserved and integrated with the context. For this section, we use an old man generated by Consistory~\cite{tewel2024training}, the famous Yokozuna for a large body, and a Dalle-3~\cite{betker2023improving} generated man for skinny body type.  In~\cref{fig:bodyshapes}, we can see that the body types are preserved.  For the second column of the men holding roses, we can see through the gap between the arm and the body for the Dalle-3 man, while Yokozuna completely blocks the background.

\begin{figure*}[t]
  \centering
  \includegraphics[width=.95\textwidth, trim=0 500 1050 50, clip]{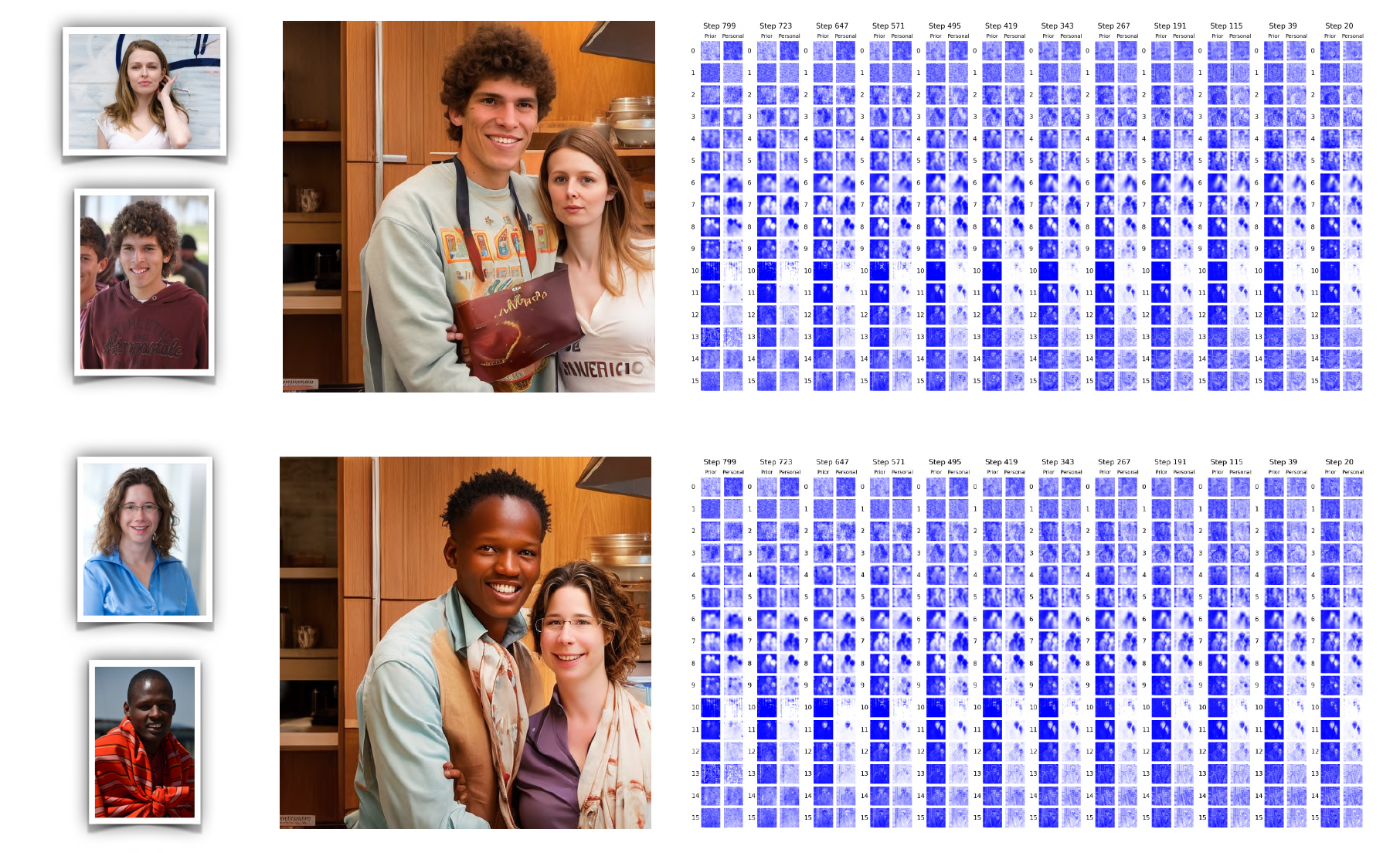}
  \caption{ \textbf{Router visualization.} The 16 rows in the visualization correspond to the 16 layers in the U-Net.
  }
  \label{fig:router-viz-old}
\end{figure*}

\section{Quantitative Results \& Analysis}

\paragraph{Evaluation metrics.}
The primary quantitative metrics we use are identity preservation (IP), and prompt consistency (PC).  
To assess IP, pairwise identity similarity is calcuated between the generated image and the input image using FaceNet~\cite{schroff2015facenet}. 
To assess PC, the average CLIP-L/14 image-text similarity is calculated following previous studies~\cite{gal2022image,xiao2023fastcomposer}.

\begin{table}    
	 
	\begin{tabular}{cc|cc}
		\bottomrule
		\multirow{2}{*}{\textbf{Methods}} & \multirow{2}{*}{\emph{OF}}  & \multicolumn{2}{c|}{\textit{\textbf{Single-Subject}}} \\
		& & IP  $\uparrow$       & PC $\uparrow$       \\
		\bottomrule
            ELITE &   $\checkmark$       & 0.228             & 0.146        \\
		Dreambooth  &               & 0.273             & 0.239       \\
	     Custom-Diffusion &       & 0.434             & 0.233   \\
        Fastcomposer &    $\checkmark$        & 0.514             & 0.243     \\
		
        Subject-Diffusion & $\checkmark$ & 0.605             & 0.228    \\
          Mixture-of-Attention &     $\checkmark$          & 0.555   & 0.202          \\
        \bottomrule
	\end{tabular} \\
 \normalsize
    \captionof{table}{Quantitative results. OF stands for ``optimization-free''.}
 \label{table:comparison}
 \end{table}

\begin{figure}
    \centering
      \centering
  \includegraphics[width=\columnwidth, trim=0 250 1000 0, clip]{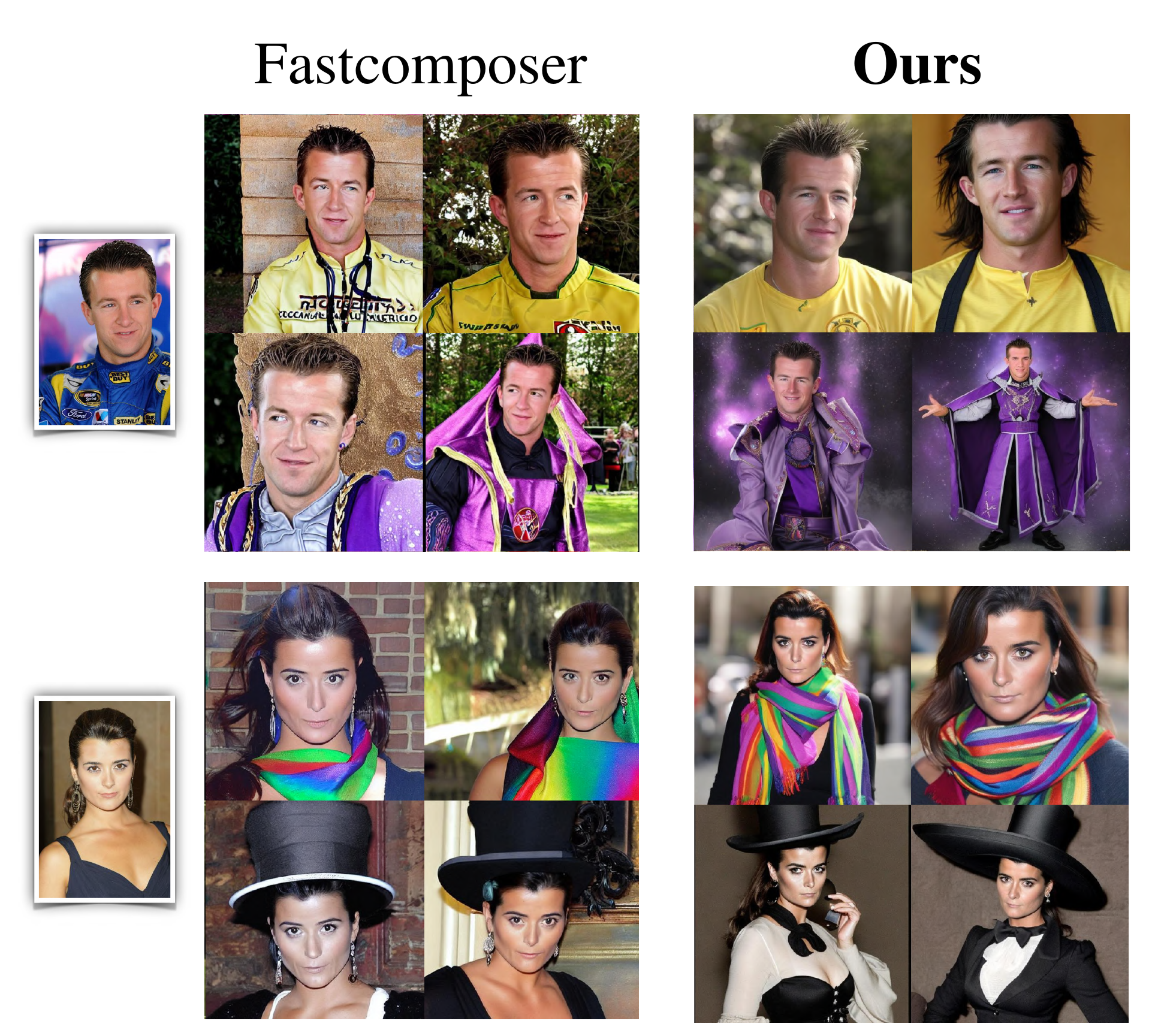}
    \captionof{figure}{Samples from the quantitative evaluation. } 
    \label{fig:quantitative-samples}
\end{figure}

We perform the same evaluation as baseline methods, and perform automated quantitative evaluation of identity preservation (IP) and prompt consistency (PC) (See~\cref{table:comparison}). 
While we perform on par with baselines like FastComposer, samples from our method have more image variation (e.g. in layout) (See~\cref{fig:quantitative-samples}).  
Also, in~\cref{fig:fastcomposer-comparison}, we can clearly see that our generated images have much better variations and interaction with the context. In the baseline, even when the text is ``riding a bike'', the bike is barely visible, and there is no clear interaction. 
However, in terms of the automated evaluation metric, having a small face region can lead to a lower score in the automated quantitative evaluation.
Note that for a fair qualitative comparison with FastComposer, we use UniPC scheduler and our prompting strategy with their checkpoint to generate baseline results.




\section{Additional Applications}
\label{sec:app:app}

\begin{figure*}[t]
  \centering
  \begin{subfigure}[t]{0.5\textwidth}
  \includegraphics[width=.95\columnwidth, trim=0 120 960 0, clip]{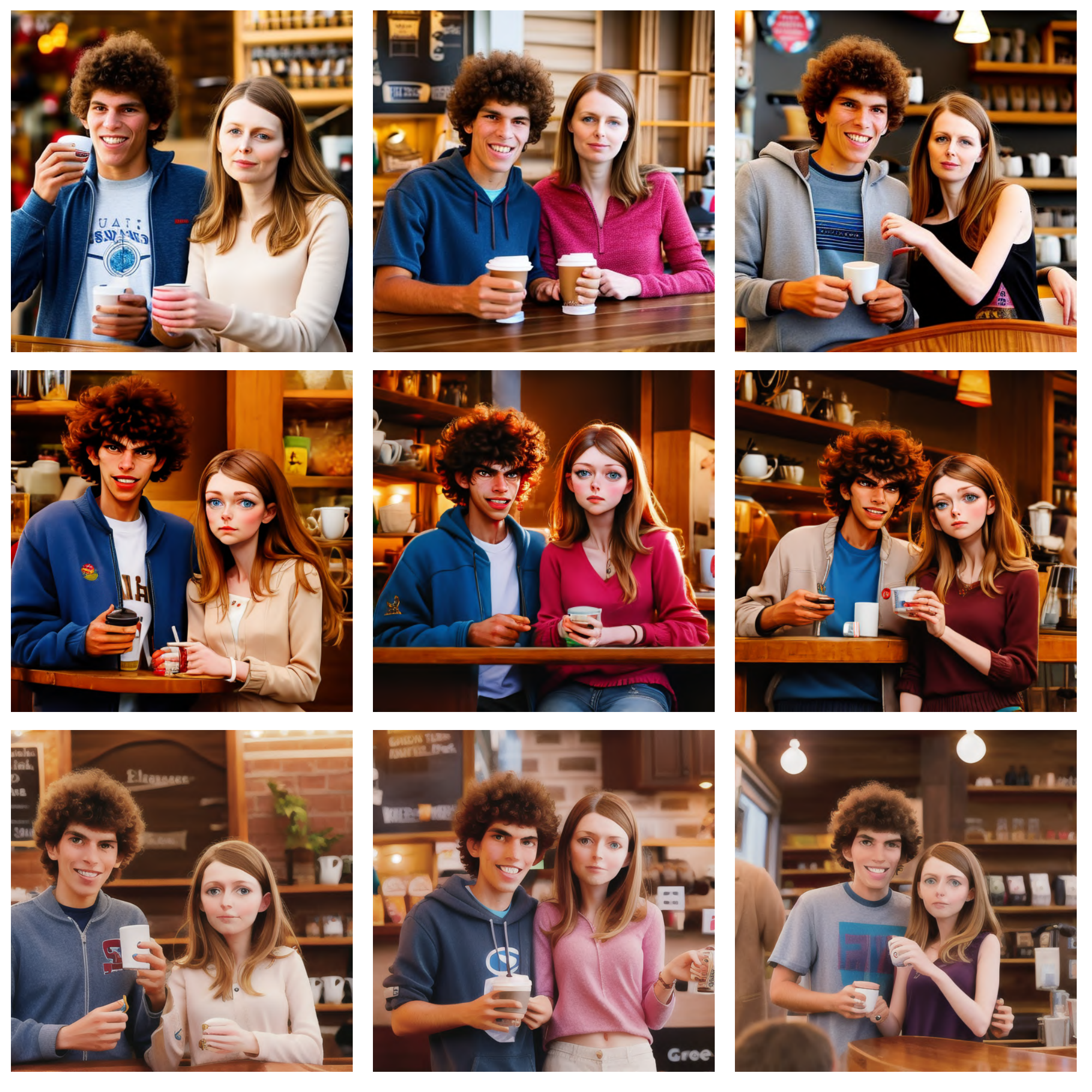}    
  \caption{Subject pair 1}
  \end{subfigure}%
  ~
  \begin{subfigure}[t]{0.5\textwidth}
  \includegraphics[width=.95\columnwidth, trim=0 120 960 0, clip]{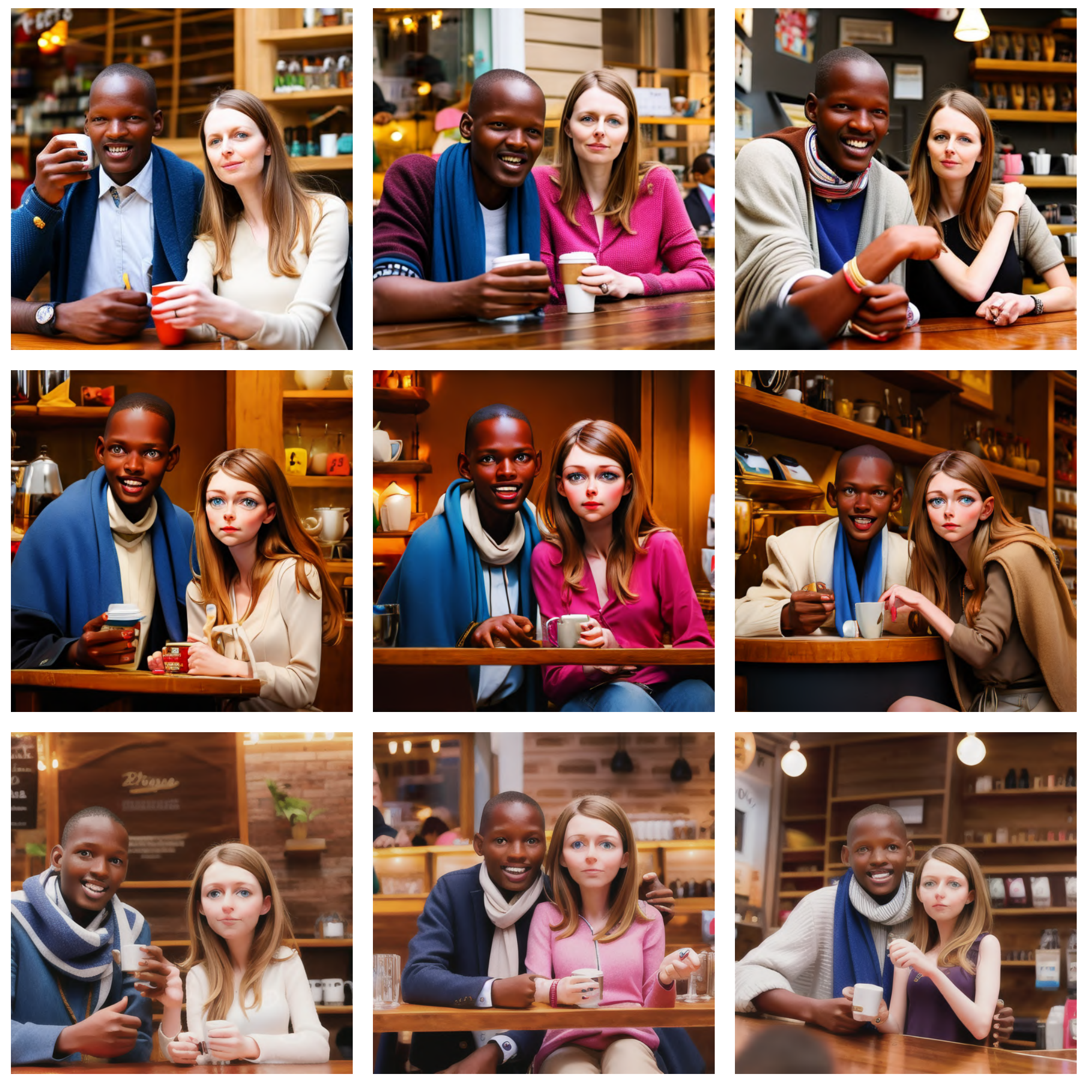}    
  \caption{Subject pair 2}
  \end{subfigure}
  
  \caption{ \textbf{Stylized generation.}  The three rows are: original MoA, $+$ ToonYou LoRA, $+$ Pixar LoRA. MoA is compatible with pretrained style LoRAs.  Adding style to MoA is as simple as loading the pretrained LoRA to the prior branch of a trained MoA during generation.
  }
  \label{fig:style}
\end{figure*}

\begin{figure}[t]
  \centering
  \includegraphics[width=\columnwidth, trim=0 100 670 0, clip]{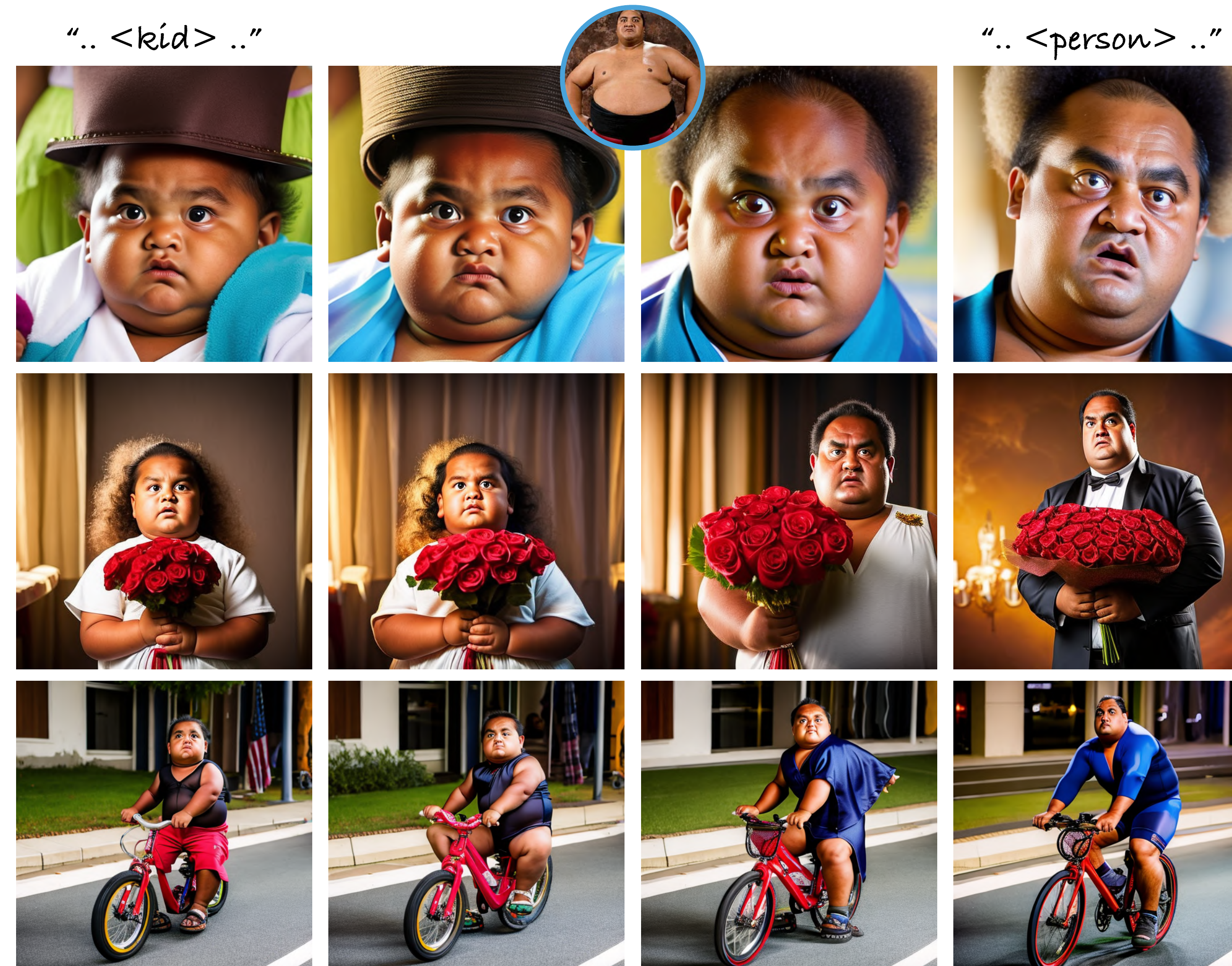}
  \caption{ \textbf{Time lapse.}  By interpolating between the token `kid' and `person', where Yokozuna's image is injected, MoA creates this time lapse sequence between Yokozuna as a kid and an adult.
  }
  \label{fig:time_lapse}
\end{figure}

\begin{figure}[t]
  \centering
  \includegraphics[width=\columnwidth]{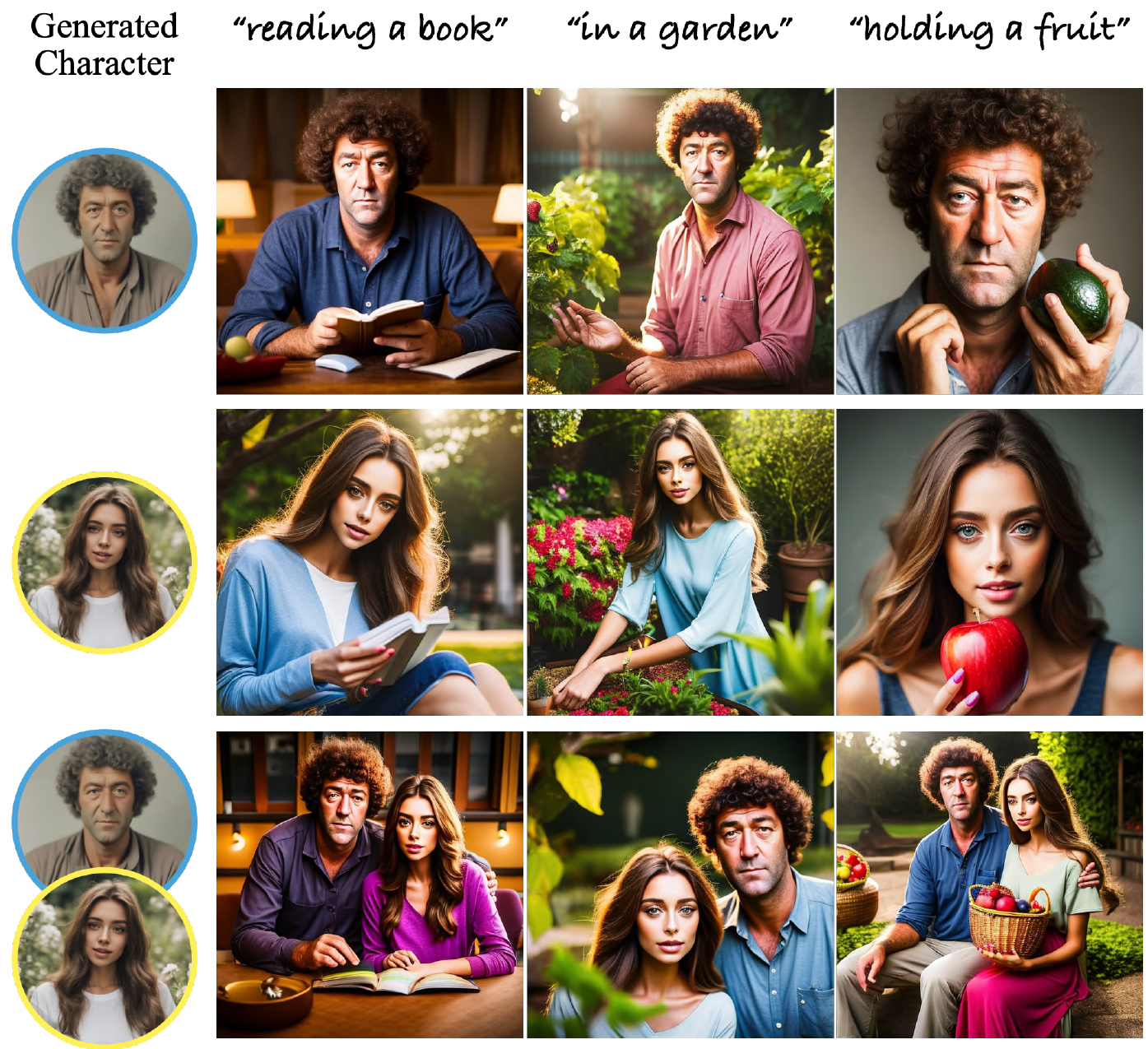}
  \caption{ \textbf{Storytelling with consistent characters.}  MoA makes it easy to put AI generated characters in new scenarios and combining different characters to form a story. 
  }
  \label{fig:story}
\end{figure}

MoA is compatible with style LoRAs (\cref{sec:app:style_swap}).
By interpolating the image and text features separately, MoA is able to generate meaningful and smooth transitions (\cref{sec:app:time_lapse}).  Lastly, the ability to generate multiple consistent characters allows creators to put AI generated characters in different scenarios and compose them to tell stories(\cref{sec:app:character}).

\subsection{Adding Style to Personalized Generation}
\label{sec:app:style_swap}

In addition to being compatible to ControlNet, the prior branch in MoA is also compatible with style LoRAs.  By combining style LoRA with MoA, users can easily generate images in different styles.  In~\cref{fig:style}, we show stylized generation using two different style LoRAs: ToonYou~\cite{toonyou}, and Pixar~\cite{pixar}.  Preserving identity across different styles/domains is a challenging task.  Identity preservation at the finest details across domain can be ill-defined.  Yet, from~\cref{fig:style}, we can clearly see the broad features of the subjects (e.g. hair style, face and body shape) are well preserved and easily recognizable.

\subsection{Time Lapse}
\label{sec:app:time_lapse}
Similar to the subject morphing by interpolating the image features, we can achieve the `time lapse' affect by interpolating between the text embeddings of `person' and `kid'.  In~\cref{fig:time_lapse}, we show images of Yokozuna at different interpolated text tokens.  
Surprisingly, MoA is able to generate Yokozuna at different ages with only a single image of him as an adult.  We hypothesize that the pretrained diffusion model has a good understanding of the visual effect of aging, and because of the strong prior preservatin of MoA, it is able to interpret the same subject at different ages.

\subsection{Storytelling with Consistent Character}
\label{sec:app:character}

With the rise of AI generated content both within the research community and the artistic community, there is significant effort put in crafting visually pleasing characters.  However, to tell a story with the generated characters consistently across different frames remains to be a challenge.  This is another application of subject-driven generation, the task we study.  In~\cref{fig:story}, we can generate consistent characters across different frames easily using our MoA model.  The man is taken from The Chosen One~\cite{avrahami2023chosen}, and the girl from the demo of IP-adapter~\cite{ye2023ip-adapter}.  Compared to The Chosen One, we can easily incorporate generated character from another method.  Compare to IP-adapter, we can easily combine the two generated characters in a single frame, which IP-adapter fails to do.

\end{document}